\documentclass[letterpaper]{article}
\usepackage{uai2020}
\usepackage[margin=1in]{geometry}

\usepackage{times}

\usepackage{microtype}
\usepackage{graphicx}
\usepackage{subfigure}
\usepackage{booktabs} % for professional tables
\usepackage{calrsfs}
\usepackage{amsmath}
\usepackage{multirow}
\usepackage{natbib}
\usepackage{amssymb}
\usepackage{mathtools}
\DeclareMathAlphabet{\pazocal}{OMS}{zplm}{m}{n}

\DeclarePairedDelimiter\floor{\lfloor}{\rfloor}

\usepackage{algpseudocode}
\usepackage{algorithm}

\usepackage{amsmath,amsfonts,bm}

\def\eqref#1{equation~\ref{#1}}
\def\floor#1{\lfloor #1 \rfloor}
\def\1{\bm{1}}

\DeclareMathAlphabet{\mathsfit}{\encodingdefault}{\sfdefault}{m}{sl}
\SetMathAlphabet{\mathsfit}{bold}{\encodingdefault}{\sfdefault}{bx}{n}

\setcitestyle{authoryear,round,citesep={;},aysep={,},yysep={;}}

\usepackage{hyperref}
\usepackage{url}

\title{Generalized Bayesian Posterior Expectation Distillation\\ for Deep Neural Networks}

\author{
{Meet P. Vadera}\textsuperscript{\rm 1}, {Brian Jalain}\textsuperscript{\rm 2}, and {Benjamin M. Marlin}\textsuperscript{\rm 1}\\
\textsuperscript{1}University of Massachusetts Amherst, \textsuperscript{2} US Army Research Laboratory\\
\texttt{\{mvadera, marlin\}@cs.umass.edu, brian.a.jalaian.civ@mail.mil}
}

\begin{document}

\maketitle

\begin{abstract}
In this paper, we present a general framework for distilling expectations with respect to the Bayesian posterior distribution of a deep neural network classifier, extending prior work on the Bayesian Dark Knowledge framework.  The proposed framework takes as input ``teacher" and student model architectures and a general posterior expectation of interest.  The distillation method performs an online compression of the selected posterior expectation using iteratively generated Monte Carlo samples. We focus on the posterior predictive distribution and expected entropy as distillation targets. We investigate several aspects of this framework including the impact of uncertainty and the choice of student model architecture. We study methods for student model architecture search from a speed-storage-accuracy perspective and evaluate down-stream tasks leveraging entropy distillation including uncertainty ranking and out-of-distribution detection.
\end{abstract}

\section{Introduction}

Deep learning models have shown promising results in the areas including computer vision, natural language processing, speech recognition, and more \citep{graves2013speech,Huang2016DenselyCC,Devlin2018BERTPO}. However, existing point estimation-based training methods for these models may result in predictive uncertainties that are not well calibrated, including the occurrence of confident errors. 

While Bayesian inference can often provide more robust posterior predictive distributions compared to point estimation-based training, the integrals required to perform Bayesian inference in neural network models are well-known to be intractable. Monte Carlo methods provide one solution to represent neural network parameter posteriors as ensembles of networks, but this requires large amounts of both storage and compute time \citep{Neal:1996:BLN:525544,welling2011bayesian}. 

To help overcome these problems, \citet{balan2015bayesian} introduced a model training method referred to as \emph{Bayesian Dark Knowledge} (BDK). BDK attempts to compress (or distill) the Bayesian posterior predictive distribution induced by the full parameter posterior of a ``teacher" network (represented via a set of Mote Carlo samples) into a significantly more compact ``student" network. The major advantage of BDK is that the computational complexity of prediction at test time is drastically reduced compared to directly computing predictions via Monte Carlo averages over the set of teacher network samples (the teacher ensemble). As a result, such posterior distillation methods have the potential to be much better suited to learning models for deployment in resource constrained settings.

However, the posterior predictive distribution is not the only statistic of the posterior distribution that is of interest. Indeed, recent work including \citet{wang2018adversarial} and \citet{malinin2020ensemble} has investigated leveraging multiple statistics of ensembles (both general ensembles and Monte Carlo representations of Bayesian posteriors) for performing tasks that leverage uncertainty quantification and uncertainty decomposition including out-of-distribution detection and uncertainty-based ranking.

In this paper, we propose a Bayesian posterior distillation framework for the classification setting that generalizes the BDK approach by directly distilling general posterior expectations.
We further generalize the BDK approach by proposing methods for efficiently searching the space of speed-storage-accuracy trade-offs for the student model, enabling more fine grained control over model size, test time, speed and predictive performance. The primary empirical contributions of this work are (1) evaluating the distillation of the posterior predictive distribution and the posterior expected entropy across a range of models, data sets, and levels of uncertainty; (2) evaluating the impact of the student model architecture and architecture search methods on distillation performance; and (3) evaluating the utility of generalized expectation distillation through the study of down-stream tasks including out-of-distribution detection and uncertainty ranking that leverage entropy distillation. We show that distillation performance can be very sensitive to student model capacity and that the proposed architecture search methods effectively expose the space of speed-storage-accuracy trade-offs. We further show that our direct generalized posterior distillation framework outperforms an adaptation of the approach of \citet{malinin2020ensemble} both on terms of distillation performance and in terms of several downstream tasks that leverage uncertainty quantification.

In the next section, we present background material and related work. In Section 3, we present the proposed framework. In Section 4, we present experiments and results. Additional details regarding data sets and experiments can be found in Appendix A, with supplemental results included in Appendix B.

\section{Background and Related Work}
In this section, we present background  and related work.

\textbf{Bayesian Neural Networks:} Let $p(y|\mathbf{x}, \theta)$ represent the probability distribution induced by a deep neural network classifier over classes $y\in\pazocal{Y}=\{1,..,C\}$ given feature vectors $\mathbf{x}\in\mathbb{R}^D$. The most common way to fit a model of this type given a data set $\pazocal{D} =\{(\mathbf{x}_i,y_i)|1\leq i\leq N\}$ is to use maximum conditional likelihood estimation, or equivalently, cross entropy loss minimization (or their penalized or regularized variants). However, when the volume of labeled data is low, there can be multiple advantages to considering a full Bayesian treatment of the model. Instead of attempting to find the single (locally) optimal parameter set $\theta_*$ according to a given criterion, Bayesian inference uses Bayes rule to define the posterior distribution $p(\theta|\pazocal{D},\theta^0)$ over the unknown parameters $\theta$ given a prior distribution $P(\theta|\theta^0)$ with prior parameters $\theta^0$ as seen in Equation \ref{eq:posterior}. 
\begin{align}
\allowdisplaybreaks[4]
\label{eq:posterior}
    p(\theta|\pazocal{D},\theta^0)& = \frac{p(\pazocal{D}|\theta)p(\theta|\theta^0)}{\int p(\pazocal{D}|\theta)p(\theta|\theta^0) d\theta}\\
\label{eq:posterior_predictive}
p(y| \mathbf{x}, \pazocal{D},\theta^0) &= \int p(y|\mathbf{x}, \theta) p(\theta|\pazocal{D},\theta^0) d\theta \notag \\
&=\mathbb{E}_{p(\theta|\pazocal{D},\theta^0)}[p(y|\mathbf{x}, \theta)]
\end{align}

\textbf{Posterior Expectations and Uncertainty Quantification:} For prediction problems in machine learning, the quantity of interest is typically not the parameter posterior itself, but the posterior predictive distribution $p(y| \mathbf{x}, \pazocal{D},\theta^0)$ obtained from it as seen in  Equation  \ref{eq:posterior_predictive}. 

However, the posterior predictive distribution is not the only statistic of the posterior distribution that is of interest. 
The decomposition of posterior uncertainty has also received recent attention in the literature. For example, \citet{Depeweg2017DecompositionOU} and \citet{malinin2020ensemble} describe the decomposition of the entropy of the posterior predictive distribution (the \emph{total uncertainty}) into \emph{expected data uncertainty} and \emph{knowledge uncertainty}. These three forms of uncertainty are related by the equation shown below:
\begin{align}
\underbrace{\pazocal{I}\left[y, \theta | \boldsymbol{x}, \pazocal{D}\right]}_{\text {Knowledge Uncertainty }} & =\underbrace{\pazocal{H}\left[\mathbb{E}_{p(\theta | \pazocal{D})}\left[p\left(y | \boldsymbol{x}, \theta\right)\right]\right]}_{\text {Total Uncertainty }} \notag \\
&-\underbrace{\mathbb{E}_{p(\theta | \pazocal{D})}\left[\pazocal{H}\left[p\left(y | \boldsymbol{x}, \theta\right)\right]\right]}_{\text {Expected Data Uncertainty }}
\end{align}
Total uncertainty, as the name suggests, measures the total uncertainty in a prediction. Expected data uncertainty measures the uncertainty arising from class overlap.   Knowledge uncertainty corresponds to the conditional mutual information between labels and model parameters and measures the disagreement between different models in the posterior. 
However, it can be efficiently computed as the difference between total uncertainty and expected data uncertainty, both of which are (functions) of posterior expectations. In recent work,   \citet{wang2018adversarial} and \citet{malinin2020ensemble} have leveraged this decomposition to explore a range of down-stream tasks that rely on uncertainty quantification and decomposition.

\label{sec:ref:approx}
\textbf{Approximate Inference Methods for Bayesian Neural Networks:} The primary problem with applying Bayesian inference to neural network models is that the distributions $p(\theta|\pazocal{D},\theta^0)$ and $p(y| \mathbf{x}, \pazocal{D},\theta^0)$ are not available in closed form, so approximations are required. We briefly review Bayesian inference approximations including variational inference (VI) \citep{jordan1999introduction} and Markov Chain Monte Carlo (MCMC) methods \citep{Neal:1996:BLN:525544,welling2011bayesian}.

In VI, an auxiliary distribution $q_{\phi} (\theta)$  is defined to approximate the true parameter posterior $p(\theta | \pazocal{D},\theta^0)$. The variational parameters $\phi$ are selected to minimize the Kullback-Leibler (KL) divergence between $q_{\phi} (\theta)$ and $p(\theta | \pazocal{D},\theta^0)$. \citet{hinton1993keeping} first studied applying VI to neural networks. \citet{graves2011practical} later presented a method based on stochastic VI with improved scalability. In the closely related family of 
expectation propagation (EP) methods \citep{minka2001expectation}, 
%The main difference between VI and EP is that in VI, we minimize $\textrm{KL}(q_{\phi} (\theta) || p(\theta | \pazocal{D},\theta^0))$, while in EP we minimize $\textrm{KL}(p(\theta | \pazocal{D},\theta^0) || q_{\phi} (\theta))$.
\citet{soudry2014expectation} present an online EP algorithm for neural networks with the flexibility of representing both continuous and discrete weights. \citet{hernandez2015probabilistic}  present the probabilistic backpropagation (PBP) algorithm for approximate Bayesian learning of neural network models, which is an example of an assumed density filtering (ADF) algorithm that, like VI and EP, generally relies on simplified posterior densities.
%
%To deal with the issues related to using simplified posterior densities (typically diagonal Gaussian) that arise in VB, EP, and PBP, 
%\citet{louizos2017multiplicative} introduced the concept of \emph{Multiplicative Normalizing Flows}. Under Multiplicative Normalizing Flows, a chain of transformations are applied to parameters sampled from the approximate posterior $q_{\phi}(\theta)$ using some additional learnable parameters $\omega$. This technique helps to create samples imitating more complex distributions than standard mean-field inference. However, this requires the flow to be invertible.

The main drawback of VB, EP, and ADF is that they typically result in biased posterior estimates for complex posterior distributions. MCMC methods provide an alternative family of sampling-based posterior approximations that are unbiased, but are often computationally more expensive to use. MCMC methods allow for drawing a correlated sequence of samples $\theta_t \sim p(\theta|\pazocal{D},\theta^0)$ from the parameter posterior. These samples can then be used to approximate the posterior predictive distribution as a Monte Carlo average as shown in Equation \ref{eq:MC_posterior_predictive}.
\begin{align}
\allowdisplaybreaks[4]
\label{eq:MC_posterior_predictive}
p(y| \mathbf{x}, \pazocal{D},\theta^0) 
&\approx  \frac{1}{T}\sum_{t=1}^T p(y|\mathbf{x}, \theta_t)\\ 
\theta_t &\sim p(\theta|\pazocal{D},\theta^0)
\end{align}
\citet{Neal:1996:BLN:525544} addressed the problem of Bayesian inference in neural networks using Hamiltonian Monte Carlo (HMC) to provide a set of posterior samples. A bottleneck with this method is that it uses the full dataset when computing the gradient needed by HMC, which is problematic for larger data sets. While this scalability problem has largely been solved by more recent methods such as stochastic gradient Langevin dynamics (SGLD) \citep{welling2011bayesian}, the problem of needing to compute over a large set of samples when making predictions at test or deployment time remains.

\textbf{Distribution Distillation:} As noted above, MCMC-based approximations are expensive in terms of both computation and storage. Bayesian Dark Knowledge \citep{balan2015bayesian} is precisely aimed at reducing the test-time computational complexity of Monte Carlo-based approximations for neural networks. In particular, the method uses SGLD to approximate the posterior distribution using a set of posterior parameter samples. These samples can be thought of as an ensemble of neural network models with identical architectures, but different parameters. 

This posterior ensemble is used as the ``teacher" in a distillation process that trains a single ``student" model to match the teacher ensemble's posterior predictive distribution \citep{hinton2015distilling}. The  major advantage of this approach is that it can drastically reduce the test time computational complexity of posterior predictive inference relative to using a Monte Carlo average computed using many samples. A shortcoming of this approach is that it only distills the posterior predictive distribution, and thus, we lose access to other posterior statistics. 

Ensemble distribution distillation (EnD\textsuperscript{2}) is a closely related approach that aims to distill the collective outputs of the models in an ensemble into a neural network that predicts the parameters of a Dirichlet distribution \citep{malinin2020ensemble}. The goal is to preserve information about distribution of outputs of the ensemble in such a way that multiple statistics of the ensemble's outputs can be efficiently approximated. The goal in this paper is broadly similar although we focus specifically on distilling much larger Monte Carlo posterior ensembles and we avoid the parametric distribution assumptions of \citep{malinin2020ensemble} by directly distilling posterior expectations of interest. 

Finally, we note that with the advent of \emph{Generative Adversarial Networks} \citep{goodfellow2014generative}, there has also been work on generative models for approximating posterior sampling. \citet{wang2018adversarial}  and \citet{henning2018approximating} both propose methods for learning to generate samples that mimic those produced by SGLD. However, while these approaches may provide a speed-up relative to running SGLD itself, the resulting samples must still be used in a Monte Carlo average to compute a posterior predictive distribution in the case of Bayesian neural networks. This is again a potentially costly operation and is exactly the computation that distillation-based methods seek to accelerate. 

\textbf{Model Compression and Pruning:} As noted above, the problem that Bayesian Dark Knowledge attempts to solve is reducing the test-time computational complexity of using a Monte-Carlo posterior to make predictions. In this work, we are particularly concerned with the issue of enabling test-time speed-storage-accuracy trade-offs. The relevant background material includes methods for network compression and pruning. 

Previous work has shown that  over-parameterized deep learning models tend to show much better learnability. Further, it has also been shown that such over-parameterized models rarely use their full capacity and can often be pruned back substatially without  significant loss of generality. 
\citet{Hassibi1993OptimalBS} use the second order derivatives of the objective function to guide pruning network connections. More recently, \citet{Han2015LearningBW} introduced a weight magnitude-based  technique for pruning connections in deep neural networks using simple thresholding.  \citet{Guo2016DynamicNS,Jin2016TrainingSD,Han2016DSDRD} introduce thresholding methods which also support restoration of connections.

A related line of work includes pruning neurons/channels/filters instead of individual weights. Pruning these components explicitly reduces the number of computations by making the networks smaller. Group LASSO-based methods have the advantage of turning the pruning problem into a continuous optimization problem with a sparsity-inducing regularizer. \citet{Zhang2018LEARNINGSS,alvarez2016learning,wen2016learning,He2017ChannelPF} are some examples that use Group LASSO regularization at their core. Similarly \citet{Louizos2017BayesianCF} use hierarchical priors to prune neurons instead of weights. An advantage of these methods over  connection-based sparsity methods is that they directly produce smaller networks after pruning. 
%(e.g., fewer units or channels) as opposed to networks with sparse weight matrices. This makes it easier to realize the resulting computational savings, even on platforms that do not directly support sparse matrix operations. 

\section{Proposed Framework}

In this section, we describe our proposed framework.

\subsection{Generalized Posterior Expectations}
As described in the previous section, different statistics derived from the posterior distribution $p(\theta|\pazocal{D},\theta^0)$ may be useful in different data analysis tasks. 
We consider the general case of inferences that take the form of posterior expectations as shown in Equation \ref{eq:posterior_expectation} where $g(y,\mathbf{x},\theta)$ is an arbitrary function of $y$, $\mathbf{x}$ and $\theta$. 
\begin{align}
\allowdisplaybreaks[4]
\label{eq:posterior_expectation}
    \mathbb{E}_{p(\theta|\pazocal{D},\theta^0)}[g(y,\mathbf{x},\theta)]&=\int  p(\theta|\pazocal{D},\theta^0)g(y,\mathbf{x},\theta) d\theta
\end{align}

Important examples of functions $g(y,\mathbf{x},\theta)$ include $g(y,\mathbf{x},\theta)=p(y|\mathbf{x},\theta)$, which results in the posterior predictive distribution $p(y| \mathbf{x}, \pazocal{D},\theta^0)$ as used in Bayesian Dark Knowledge. The choice $g(y,\mathbf{x},\theta)=\sum_{y'=1}^C p(y'|\mathbf{x},\theta)\log p(y'|\mathbf{x},\theta)$ yields the expected data uncertainty introduced in the previous section.  The choice $g(y,\mathbf{x},\theta)=p(y|\mathbf{x},\theta)(1-p(y|\mathbf{x},\theta))$ results in the posterior marginal variance of class $y$ given $\mathbf{x}$. We use the posterior predictive distribution and expected data uncertainty as examples throughout this work.

\subsection{Generalized Posterior Expectation Distillation}

Our goal is to learn to approximate posterior expectations $\mathbb{E}_{p(\theta|\pazocal{D},\theta^0)}[g(y,\mathbf{x},\theta)]$ under a given teacher model architecture using a given student model architecture. The method that we propose takes as input the teacher model  $p(y|\mathbf{x},\theta)$, the prior $p(\theta|\theta^0)$, a labeled data set $\pazocal{D}$, an unlabeled data set $\pazocal{D}'$, the function $g(y,\mathbf{x},\theta)$, a student model $f(y,\mathbf{x}|\phi)$, an expectation estimator, and a loss function $\ell(\cdot,\cdot)$ that measures the error of the approximation given by the student model $f(y,\mathbf{x}|\phi)$.  Similar to \cite{balan2015bayesian}, we propose an online distillation method based on the use of the SGLD sampler. We describe all of the components of the framework in the sections below, and provide a complete description of the resulting method in Algorithm \ref{algorithm:ped}.

% \begin{algorithm*}[t]\caption{Generalized Posterior Expectation Distillation}
% \label{algorithm:ped}
% \begin{algorithmic}[1]
% \Procedure{GPED}{$\pazocal{D},\pazocal{D}', p(y|\mathbf{x},\theta), \theta^0, g, f, U,\ell, R, M,M',H,B, \lambda, \{\eta_t\}_{t=1}^T, \{\alpha_s\}_{s=1}^S$ }
% \State Initialize $s=0$, $\phi_0$, $\theta_0$, $\hat{g}_{yi0}=0$, $m_{i0}=0$,$\eta_0$
% \For{$t=0$ to $T$}
%   \State Sample $\pazocal{S}$ from $\pazocal{D}$ with $|\pazocal{S}|=M$
%   \State $\theta_{t+1} \!\leftarrow\theta_{t}+ \!\frac{\eta_{t}}{2}\!\left(\!\nabla_{\theta} \log p(\theta | \theta^0)\!+\!\frac{N}{M} \sum_{i \in \pazocal{S}} \nabla_{\theta} \log p\left(y_{i} | x_{i}, \theta_{t}\right)\!\right)+ z_{t}$
%   \If{$\mod(t,H)=0$ and $t>B$}
%     \State Sample $\pazocal{S}'$ from $\pazocal{D}'$ with $|\pazocal{S}'|=M'$
%     \For{$i \in \pazocal{S}'$}
%         \State $\hat{g}_{yis+1} \leftarrow U(\hat{g}_{yis}, \theta_t, m_{is})$
%       \State $m_{is+1} \leftarrow m_{is}+1$
%     \EndFor
%     \State $\phi_{s+1} \leftarrow \phi_{s} + \alpha_s \left(\frac{N'}{M'}\sum_{i\in \pazocal{S}'}\sum_{y\in\pazocal{Y}}\nabla_{\phi}\ell\big(\hat{g}_{yis+1}, f(y,\mathbf{x}_i|\phi_s)\big) + \lambda \nabla_{\phi}R(\phi_s) \right)$
%     \State $s\leftarrow s+1$
%   \EndIf
% \EndFor
% \EndProcedure
% \end{algorithmic}
% \end{algorithm*}

\begin{algorithm}[t]\caption{Generalized Posterior Expectation Distillation}
\label{algorithm:ped}
\hspace*{\algorithmicindent} \textbf{Input}: $\pazocal{D}$, $\pazocal{D}'$, $ p(y|\mathbf{x}$,$\theta)$, $\theta^0$, $g$, $f$, $U$, $\ell$, $R$, $M$, $M'$, $H$, $B$, $ \lambda$, $\{\eta_t\}_{t=1}^T$, $\{\alpha_s\}_{s=1}^S$
\begin{algorithmic}[1]
\Procedure{GPED}{}
\State Initialize $s=0$, $\phi_0$, $\theta_0$, $\hat{g}_{yi0}=0$, $m_{i0}=0$,$\eta_0$
\For{$t=0$ to $T$}
  \State Sample $\pazocal{S}$ from $\pazocal{D}$ with $|\pazocal{S}|=M$
%   \State $\theta_{t+1} \!\leftarrow\theta_{t}+ \!\frac{\eta_{t}}{2}\!\left(\!\nabla_{\theta} \log p(\theta | \theta^0)\!+\!\frac{N}{M} \sum_{i \in \pazocal{S}} \nabla_{\theta} \log p\left(y_{i} | x_{i}, \theta_{t}\right)\!\right)+ z_{t}$
  \State $\theta_{t+1} \!\leftarrow\theta_{t} + \!\frac{\eta_{t}}{2}\!\Bigl(\!\nabla_{\theta} \log p(\theta | \theta^0) 
 +\!\frac{N}{M} \sum_{i \in \pazocal{S}} \nabla_{\theta} \log p\left(y_{i} | x_{i}, \theta_{t}\right)\!\Bigr) + z_{t}$
  \If{$\mod(t,H)=0$ and $t>B$}
    \State Sample $\pazocal{S}'$ from $\pazocal{D}'$ with $|\pazocal{S}'|=M'$
    \For{$i \in \pazocal{S}'$}
        \State $\hat{g}_{yis+1} \leftarrow U(\hat{g}_{yis}, \theta_t, m_{is})$
      \State $m_{is+1} \leftarrow m_{is}+1$
    \EndFor
    \State $\phi_{s+1} \leftarrow \phi_{s} + \alpha_s \Bigl(\frac{N'}{M'}\sum_{i\in \pazocal{S}'}\sum_{y\in\pazocal{Y}}\nabla_{\phi}\ell\big(\hat{g}_{yis+1}, f(y,\mathbf{x}_i|\phi_s)\big) + \lambda  \nabla_{\phi}R(\phi_s) \Bigr) $
    \State $s\leftarrow s+1$
  \EndIf
\EndFor
\EndProcedure
\end{algorithmic}
\end{algorithm}

\textbf{SGLD Sampler:} The prior distribution over the parameters $p(\theta|\theta^0)$ is chosen to be a spherical Gaussian distribution with mean $\mu=0$ and precision $\tau$ (we thus have $\theta^0=[\mu,\tau]$). We define $\pazocal{S}$ to be a minibatch of size M drawn from $\pazocal{D}$.  $\theta_t$ denotes the parameter set sampled for the teacher model at sampling iteration $t$, while $\eta_t$ denotes the step size for the teacher model at iteration $t$. The Langevin noise is denoted by $z_t \sim \pazocal{N}(0, \eta_t I )$.  The sampling update for SGLD is given  b: $\theta_{t+1} \!\leftarrow\theta_{t} + \Delta{\theta}_t$ where $\Delta{\theta}_t$ is defined as:
{\footnotesize
\begin{align}
\Delta{\theta}_t=&  \!\frac{\eta_{t}}{2}\!\Biggl(\!\nabla_{\theta} \log p(\theta | \theta^0) 
 +\!\frac{N}{M} \sum_{i \in \pazocal{S}} \nabla_{\theta} \log p\left(y_{i} | x_{i}, \theta_{t}\right)\!\Biggr) + z_{t}
\end{align}
}
\textbf{Distillation Procedure:} For the distillation learning procedure, we make use of a secondary unlabeled data set $\pazocal{D}'=\{\mathbf{x}_i|1\leq i\leq N'\}$. This data set could use feature vectors from  the primary data set $\pazocal{D}$, or a larger data set. We note that due to autocorrelation in the sampled teacher model parameters $\theta_t$, we may not want to run a distillation update for every Monte Carlo sample drawn. We thus use two different iteration indices: $t$ for SGLD iterations and $s$ for distillation iterations. 

On every distillation step $s$, we sample a minibatch $\pazocal{S}'$ from $\pazocal{D}'$ of size $M'$. For every data case $i$ in $\pazocal{S}'$, we update an estimate $\hat{g}_{yis}$ of the posterior expectation using the most recent parameter sample $\theta_t$, obtaining an updated estimate $\hat{g}_{yis+1}\approx \mathbb{E}_{p(\theta|\pazocal{D},\theta^0)}[g(y,\mathbf{x},\theta)]$ (we discuss update schemes in the next section).  Next, we use the minibatch of examples $\pazocal{S}'$ to update the student model. To do so, we take a step $\phi_{s+1} \leftarrow \phi_{s} + \alpha_s\Delta^{\phi}_t$ in the gradient direction of the regularized empirical risk of the student model as shown below where  $\alpha_s$ is the student model learning rate, $R(\phi)$ is the regularizer, and $\lambda$ is the regularization hyper-parameter. We next discuss the estimation of the expectation targets $\hat{g}_{yis}$.
{\footnotesize
\begin{align}
\Delta{\phi}_t =  \frac{N'}{M'}\sum_{i\in \pazocal{S}'}\sum_{y\in\pazocal{Y}}\nabla_{\phi}\ell\big(\hat{g}_{yis+1}, f(y,\mathbf{x}_i|\phi_s)\big) 
  + \lambda  \nabla_{\phi}R(\phi_s)  
\end{align}
}

\textbf{Expectation Estimation:} Given an explicit collection of posterior samples $\theta_1,...,\theta_s$, the standard Monte Carlo estimate of $\mathbb{E}_{p(\theta|\pazocal{D},\theta^0)}[g(y,\mathbf{x},\theta)]$ is simply $\hat{g}_{yis} = (1/S)\sum_{j=1}^s g(y,\mathbf{x}_i,\theta_j)$. However, this estimator requires retaining the sequence of samples $\theta_1,...,\theta_s$, which may not be  feasible in terms of storage cost. Instead, we consider the application of an online update function. We define $m_{is}$ to be the count of the number of times data case $i$ has been sampled up to and including distillation iteration $s$. An online update function $U(\hat{g}_{yis}, \theta_t, m_{is})$ takes as input the current estimate of the expectation, the current sample of the model parameters, and the number of times data case $i$ has been sampled,  and produces an updated estimate of the expectation $\hat{g}_{yis+1}$. 
Below, we define two different versions of the function. $U_s(\hat{g}_{yis}, \theta_t, m_{is})$, updates $\hat{g}_{yis}$ using the current sample only, while $U_o(\hat{g}_{yis}, \theta_t, m_{is})$ performs an online update equivalent to a full Monte Carlo average. 
{\footnotesize
\begin{align}
U_s(\hat{g}_{yis}, \theta_t, m_{is}) &=
g(y,\mathbf{x}_i,\theta_t)\\
U_o(\hat{g}_{yis}, \theta_t, m_{is}) &=
\frac{1}{m_{is+1}}\big(
m_{is}\cdot\hat{g}_{yis}
+ g(y,\mathbf{x}_i,\theta_t)\big)
\end{align}
}
We note that both update functions provide unbiased estimates of $\mathbb{E}_{p(\theta|\pazocal{D},\theta^0)}[g(y,\mathbf{x},\theta)]$ after a suitable burn-in time $B$. The online update $U_o(.)$ will generally result in lower variance in the estimated values of $\hat{g}_{yis}$, but it comes at the cost of needing to explicitly maintain the expectation estimates $\hat{g}_{yis}$ across learning iterations, increasing the storage cost of the algorithm. It is worthwhile noting that the extra storage and computation cost required by $U_o$ grows linearly in the size of the training set for the student. By contrast, the fully stochastic update is memoryless in terms of past expectation estimates, so the estimated expectations  $\hat{g}_{yis}$ do not need to be retained across iterations resulting in a substantial space savings.  

\textbf{General Algorithm and Special Cases:} We show a complete description of the proposed method in Algorithm \ref{algorithm:ped}. The algorithm takes as input the teacher model  $p(y|\mathbf{x},\theta)$, the parameters of the prior $P(\theta|\theta^0)$, a labeled data set $\pazocal{D}$, an unlabeled data set $\pazocal{D}'$, the function $g(y,\mathbf{x},\theta)$, the student model $f(y,\mathbf{x}|\phi)$, an online expectation estimator $U(\hat{g}_{yis}, \theta_t, m_{is})$, a loss function $\ell(\cdot,\cdot)$ that measures the error of the approximation given by $f(y,\mathbf{x}|\phi)$, a regularization function $R()$ and regularization hyper-parameter $\lambda$, minibatch sizes $M$ and $M'$, the thinning interval parameter $H$, the SGLD burn-in time parameter $B$ and step size schedules for the step sizes $\eta_t$ and $\alpha_s$.

We note that the original Bayesian Dark Knowledge method is recoverable as a special case of this framework via the the choices $g(y,\mathbf{x},\theta)=p(y|\mathbf{x},\theta)$, $\ell(p,q)=-p\log(q)$, $U = U_s$ and $p(y|\mathbf{x},\theta)=f(y,\mathbf{x},\phi)$ (e.g., the architecture of the student is selected to match that of the teacher). The original approach also uses a distillation data set $\pazocal{D}'$ obtained from $\pazocal{D}$ by adding randomly generated noise to instances from $\pazocal{D}$ on each distillation iteration, taking advantage of the fact that the choice $U = U_s$ means that no aspect of the algorithm scales with $|\pazocal{D}'|$. 

Our general framework allows for other trade-offs, including reducing the variance in the estimates of $\hat{g}_{yis}$ at the cost of additional storage in proportion to $|\pazocal{D}'|$. We also note that the loss function $\ell(p,q)=-p\log(q)$ and the choice $g(y,\mathbf{x},\theta)=p(y|\mathbf{x},\theta)$ are somewhat of a special case when used together as even when the full stochastic expectation update $U_s$ is used, the resulting distillation parameter gradient is unbiased. To distill posterior expected entropy (e.g., expected data uncertainty), we set $g(y,\mathbf{x},\theta)=\sum_{y\in\pazocal{Y}}p(y|\mathbf{x},\theta)\log p(y|\mathbf{x},\theta)$, $U = U_o$ and $\ell(h,h') = |h-h'|$. 

\subsection{Model Compression and Pruning} 

One of the primary motivations for the original Bayesian Dark Knowledge approach is that it provides an approximate inference framework that results in significant computational and storage savings at test time. However, a drawback of the original approach is that the architecture of the student is chosen to match that of the teacher. As we will show in Section 4, this will sometimes result in a student network that has too little capacity. On the other hand, if we plan to deploy the student model in a low resource compute environment, the teacher architecture may not meet the specified computational constraints. In either case, we need a general approach for selecting an architecture for the student model.

To begin to explore this problem, we consider two basic approaches to choosing student model architectures that enable trading off test time inference speed and storage for accuracy (or more generally, lower distillation loss). A helpful aspect of the distillation process relative to a de novo architecture search problem is that the architecture of the teacher model is available as a starting point.  As a first approach, we consider wrapping the proposed GPED algorithm with an explicit search over a set of student models that are ``close" to the teacher. Specifically, we consider a search space obtained by starting from the teacher model and applying a width multiplier to the width of every fully connected layer and a kernel multiplier to the number of kernels in every convolutional layer. While this search requires exponential time in the number of layers, it provides a baseline for evaluating other methods.

As an alternative approach with better computational complexity, we leverage the regularization function $R(\phi)$ included in the GPED framework to prune a large initial network using group $\ell_1/\ell_2$ regularization \citep{Zhang2018LEARNINGSS,wen2016learning}. To apply this approach, we first must partition the parameters in the parameter vector $\phi$ across $K$ groups $\pazocal{G}_k$. The form of the regularizer is 
$R(\phi) = \sum_{k=1}^K  \big(\textstyle \sum_{j\in \pazocal{G}_k } \phi_j^2 \big)^{1/2}$.
As is well-established in the literature, this regularizer causes all parameters in a group to go to zero simultaneously when they are not needed in a model. To use it for model pruning for a unit in a fully connected layer, we collect all of that unit's inputs into a group. Similarly, we collect all of the incoming weights for a particular channel in a convolution layer together into a group. If all incoming weights associated with a unit or a channel have magnitude below a small threshold $\epsilon$, we can explicitly remove them from the model, obtaining a more compact architecture. We also fine-tune our models after pruning. 

Finally, we note that any number of weight compressing, pruning, and architecture search methods could be combined with the GPED framework. Our goal is not to exhaustively compare such methods, but rather to demonstrate that GPED is sensitive to the choice of student model to highlight the need for additional research on the problem of selecting student model architectures.

\section{Experiments and Results}
In this section, we present experiments and results evaluating the proposed approach using multiple data sets, posterior expectations, teacher model architectures, student model architectures, basic architecture search methods, and multiple down-stream tasks. We begin by providing an overview of the experimental protocols used.

\subsection{Experimental Protocols}
\textbf{Data Sets:} We use the MNIST  \citep{lecun1998mnist} and CIFAR10 \citep{krizhevsky2009learning} data sets as base data  sets in our experiments. In the case of MNIST, posterior predictive uncertainty is very low, so we introduce two different modifications to explore the impact of uncertainty on distillation performance. The first modification is simply to subsample the data. The second modification is to introduce occlusions into the data set using randomly positioned square masks of different sizes, resulting in masking rates from $0\%$ to $86.2\%$. For CIFAR10, we only use sub-sampling. Full details for both data sets and the manipulations applied can be found in Appendix \ref{appendix:datasets}.

\textbf{Models:} We evaluate a total of three teacher models in this work: a three-layer fully connected network (FCNN) for MNIST matching the architecture used by \cite{balan2015bayesian}, a four-layer convolutional network for MNIST, and a five-layer convolutional network for CIFAR10. Full details of the teacher model architectures are given in Appendix \ref{appendix:models}. For exhaustive search for student model architectures, we use the teacher model architectures as base models and search over a space of layer width multipliers $K_1$ and $K_2$ that can be used to expand sets of layers in the teacher models. A full description of the search space of student models can be found in Appendix \ref{appendix:models}.  

\textbf{Distillation Procedures:} We consider distilling both the posterior predictive distribution and the posterior entropy, as described in the previous section. For the posterior predictive distribution, we use the stochastic expectation estimator $U_s$ while for entropy we experiment with both estimators. We allow $B=1,000$ burn-in iterations for MNIST and $B=10,000$ for CIFAR10, and total of $T= 10^6$ training iterations. The prior hyper-parameters, learning rate schedules and other parameters vary by data set or distillation target and are fully described in Appendix \ref{appendix:models}.

\begin{table}[t]
    \centering
    \caption{Results of posterior distillation when the student architecture is fixed to match the teacher architecture and base data sets are used with no sub-sampling or occlusion.}
    \resizebox{\linewidth}{!}{
    \begin{tabular}{cccc}
    \toprule
       \begin{tabular}[c]{@{}c@{}}Model \& \\ Dataset\end{tabular}  &   \begin{tabular}[c]{@{}c@{}}Teacher\\ NLL\end{tabular} & \begin{tabular}[c]{@{}c@{}}Student\\ NLL\end{tabular} &\begin{tabular}[c]{@{}c@{}}MAE\\ (Entropy)\end{tabular}\\
    \midrule
        FCNN - MNIST & 0.052 & 0.082 & 0.016\\
        CNN - MNIST & 0.022 & 0.053 & 0.016\\
        CNN - CIFAR10 & 0.671 & 0.932 & 0.245\\
    \bottomrule
    \end{tabular}
    }
    
    \label{tab:main:exp1}
\end{table}

\begin{figure*}[t]
    \centering
    \subfigure[]{\includegraphics[width=0.31\textwidth]{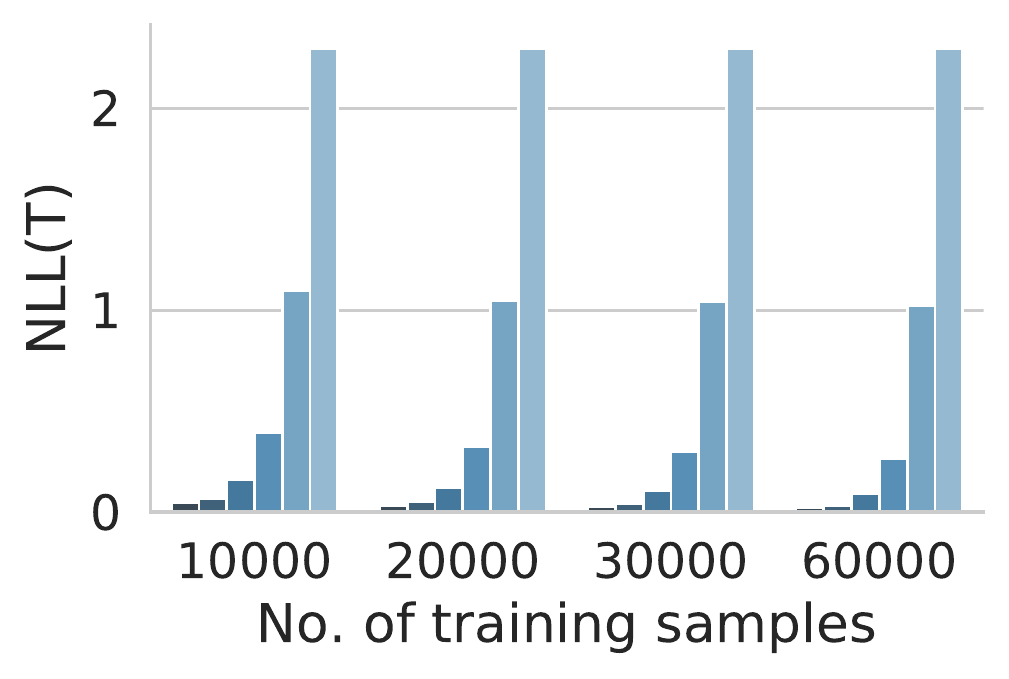}}
    \subfigure[]{\includegraphics[width=0.31\textwidth]{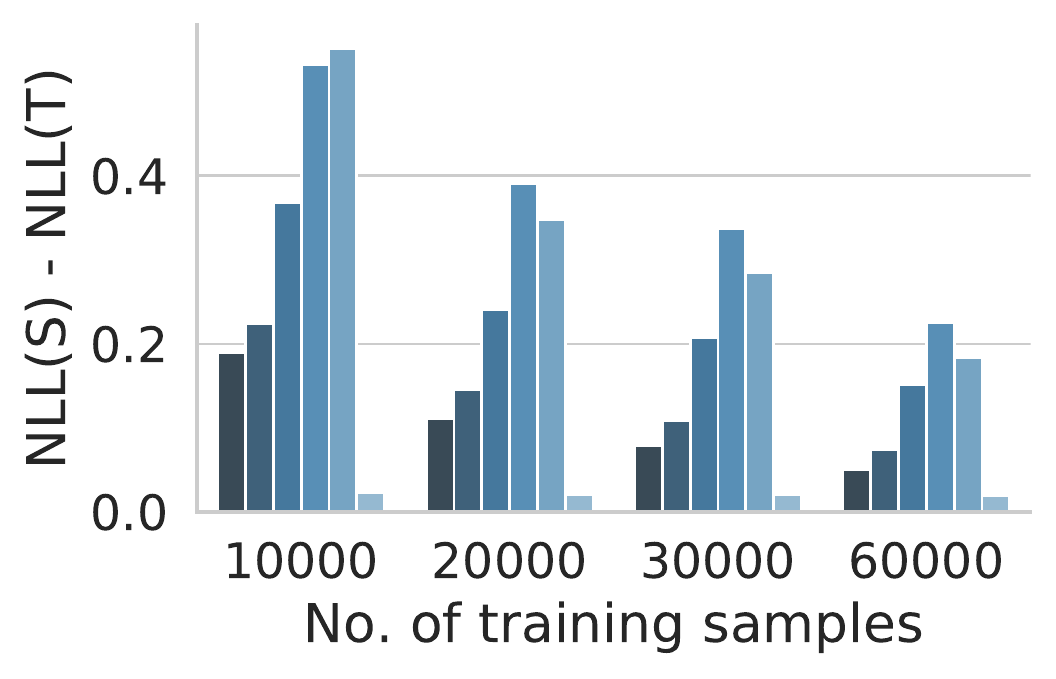}}
    \subfigure[]{\includegraphics[width=0.36\textwidth]{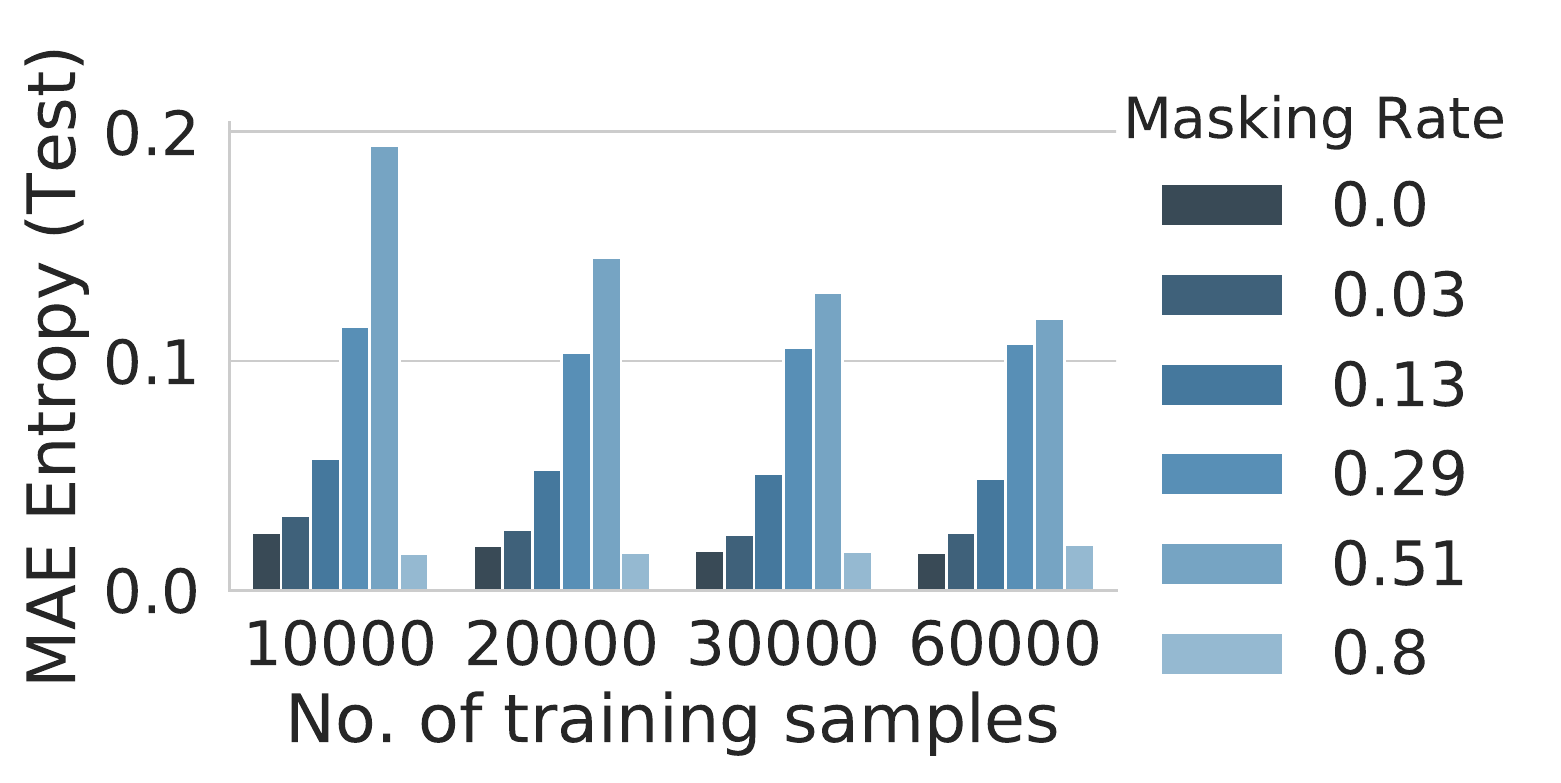}}
    \vspace{-1.em}
%
%
%\begin{figure}[htbp]
%    \centering
    % \subfigure[]{\includegraphics[width=0.33\textwidth]{figures/cifar_cnn/cifar_cnn_masking_teacher_nll_results.pdf}}
    % \subfigure[]{\includegraphics[width=0.33\textwidth]{figures/cifar_cnn/cifar_cnn_masking_delta_nll_results.pdf}
    % }
    % \subfigure[]{\includegraphics[width=0.32\textwidth]{figures/cifar_cnn/cifar_cnn_masking_delta_entropy_results.pdf}}
    \caption{Distillation performance using CNNs on MNIST while varying data set size and masking rate. (a) Test negative log likelihood of the teacher posterior predictive distribution. (b) Difference in test negative log likelihood between student and teacher posterior predictive distribution estimates. (c) Difference between teacher and student posterior entropy estimates on test data set.
    % Bottom Row: Distillation performance using CNNs on CIFAR10 while varying data set size. (d) Test negative log likelihood of the teacher posterior predictive distribution. (e) Difference in test negative log likelihood between student and teacher posterior predictive distribution estimates. (f) Difference between teacher and student posterior entropy estimates on test data set. In the plots above, S denotes the student and T denotes the teacher.
    }
    \label{fig:main:cnn_mnist_robustness_results}
    % \label{fig:main:cnn_cifar_robustness_results}

\end{figure*} 

\subsection{Experiments}

\textbf{Experiment 1: Distilling Posterior Expectations} For this experiment, we use the MNIST and CIFAR10 datasets without any subsampling or masking. For each dataset and model, we consider separately distilling the posterior predictive distribution and the posterior entropy. We fix the architecture of the student to match that of the teacher. To evaluate the performance while distilling the posterior predictive distribution, we use the negative log-likelihood (NLL) of the model on the test set. For evaluating the performance of distilling posterior entropy, we use the mean absolute difference between the teacher ensemble's entropy estimate and the student model output on the test set. The results are given in Table \ref{tab:main:exp1}. First, we note that the FCNN NLL results on MNIST closely replicate the results in \cite{balan2015bayesian}, as expected. We also note that the error in the entropy is low for both the FCNN and CNN architectures on MNIST. However, the student model fails to match the NLL of the teacher on CIFAR10 and the entropy MAE is also relatively high. In Experiment 2, we will investigate the effect of increasing uncertainty , while in Experiment 3 we will investigate the impact of student architectures.

\textbf{Experiment 2: Robustness to Uncertainty} We build on Experiment 1 by exploring methods for increasing posterior uncertainty on MNIST (sub-sampling and masking) and CIFAR10 (sub-sampling). We consider the cross product of four sub-sampling rates and six masking rates for MNIST and three sub-sampling rates for CIFAR10. We consider the posterior predictive distribution and posterior entropy distillation targets. For the  posterior predictive distribution we report the negative log likelihood (NLL) of the teacher, and the NLL gap between the teacher and student. For entropy, we report the mean absolute error between the teacher ensemble and the student. All metrics are evaluated on held-out test data. We also restrict the experiment to the case where the student architecture matches the teacher architecture, mirroring the Bayesian Dark Knowledge approach.
In Figure \ref{fig:main:cnn_mnist_robustness_results}, we show the results for the convolutional models on MNIST. The FCNN results are similar to the CNN results on MNIST and are shown in Figure \ref{fig:fcnn_mnist_robustness_results} along with the CNN results on CIFAR10 in Figure \ref{fig:cnn_cifar_robustness_results} in Appendix \ref{appendix:experiments}. In Appendix \ref{appendix:experiments}, we also provide a performance comparison between the $U_o$ and $U_s$ estimators while distilling posterior expectations.

As expected, the NLL of the teacher decreases as the data set size increases. We observe that changing the number of training samples has a similar effect on NLL gap for both CIFAR10 and MNIST. More specifically, for any fixed masking rate of MNIST (and zero masking rate for CIFAR10), we can see that the NLL difference between the student and teacher decreases with increasing training data. However, for MNIST  we can see that the teacher NLL increases much more rapidly as a function of the  masking rate. Moreover, the gap between the teacher and student peaks for moderate values of the masking rate. This fact is explained through the observation that when the masking rate is low, posterior uncertainty is low, and distillation is relatively easy. On the other hand, when the masking rate is high, the teacher essentially outputs the uniform distribution for every example, which is very easy for the student to represent. As a result, the moderate values of the masking rate result in the hardest distillation problem and thus the largest performance gap. For varying masking rates, we see exactly the same trend for the gap in posterior entropy predictions on MNIST. However, the gap for entropy prediction increases as a function of data set size for CIFAR10. Finally, as we would expect, the performance of distillation using the $U_o$ estimator is almost always better than that of the $U_s$ estimator (see Appendix \ref{appendix:experiments}).

The key finding of this experiment is that the quality of the approximations provided by the student model can significantly vary as a function of properties of the underlying data set. In the next experiment, we address the problem of searching for improved student model architectures.

\begin{figure*}[!t]
    \centering
    \subfigure[]{\includegraphics[width=0.24\textwidth]{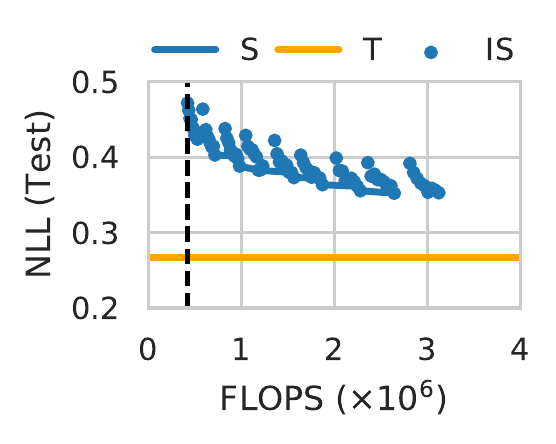}}
    \subfigure[]{\includegraphics[width=0.24\textwidth]{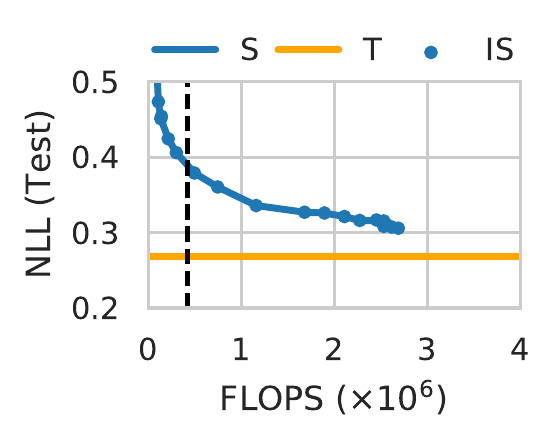}}
    \subfigure[]{\includegraphics[width=0.24\textwidth]{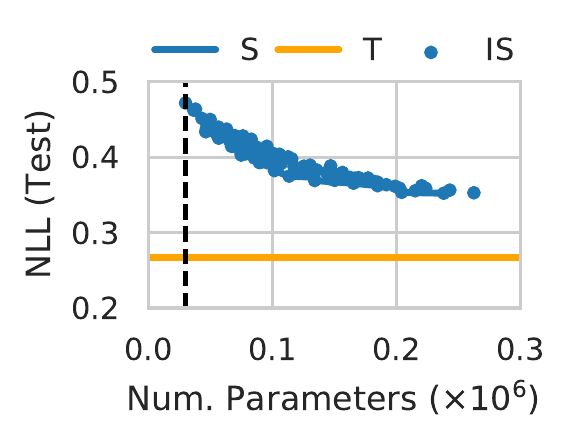}}
    \subfigure[]{\includegraphics[width=0.24\textwidth]{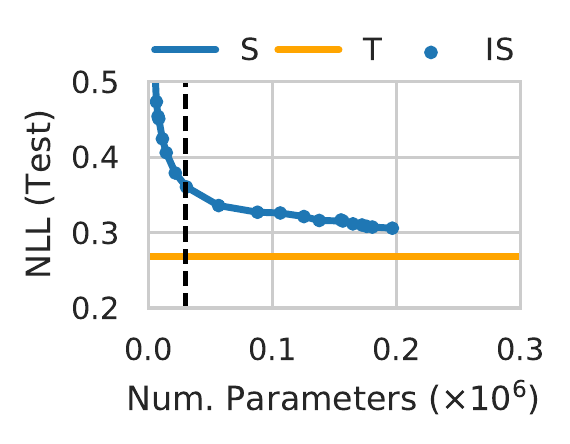}}
    \vspace{-1em}
%\end{figure}    
%
%\begin{figure}[htbp]
%    \centering
    % \subfigure[]{\includegraphics[width=0.24\textwidth]{figures/cifar_cnn/cifar_cnn_nll_flops_exhaustive_results_new.pdf}}
    % \subfigure[]{\includegraphics[width=0.24\textwidth]{figures/cifar_cnn/cifar_cnn_nll_flops_pruning_results_new.pdf}}
    % \subfigure[]{\includegraphics[width=0.24\textwidth]{figures/cifar_cnn/cifar_cnn_nll_num_params_exhaustive_results_new.pdf}}
    % \subfigure[]{\includegraphics[width=0.24\textwidth]{figures/cifar_cnn/cifar_cnn_nll_num_params_pruning_results_new.pdf}}
     \caption{
     NLL-Storage-Computation tradeoff while using CNNs on MNIST with masking rate $29\%$. Test negative log likelihood of posterior predictive distribution vs FLOPS found using (a) exhaustive search and (b) group $\ell_1/\ell_2$ with pruning. Test negative log likelihood of posterior predictive distribution vs storage found using (c) exhaustive search and (d) group $\ell_1/\ell_2$ with pruning. The optimal student model for this configuration is obtained with group $\ell_1/\ell_2$ pruning. It has approximately $6.6 \times$ the number of parameters and $6.4 \times$ the FLOPS of the base student model. Notation: ``S" - pareto frontier of the student models, ``T" - Teacher, ``IS" - Individual Student. The black dashed line denotes the FLOPS/no. of parameters of the base student model having the same architecture as a teacher model.
    %  Bottom Row:NLL-Storage-Computation tradeoff while using CNNs on CIFAR10 with training set size of 20,000 samples. (e,f) Test negative log likelihood of posterior predictive distribution vs FLOPS found using exhaustive search and group $\ell_1/\ell_2$ with pruning.  (g,h) Test negative log likelihood of posterior predictive distribution vs storage found using exhaustive search and group $\ell_1/\ell_2$ with pruning.\vspace{-1em}}
    }
    \label{fig:main:cnn_nll_tradeoff_results}
    % \label{fig:main:cifar_cnn_nll_tradeoff_results}
\end{figure*}    

\textbf{Experiment 3: Student  Model Architectures} In this experiment, we compare exhaustive search  to the group $\ell_1/\ell_2$ (group lasso) regularizer combined with pruning. For the pruning approach, we start with the largest student model considered under exhaustive search, and prune back from there using different regularization parameters $\lambda$, leading to different student model architectures.  We present results in terms of performance versus computation time (estimated in FLOPS), as well as performance vs storage cost (estimated in number of parameters). As performance measures for the posterior predictive distribution, we consider accuracy and negative log likelihood. For entropy, we use mean absolute error. In all cases, results are reported on test data. We consider both fully connected and convolutional models. 

% Figure \ref{fig:main:cnn_nll_tradeoff_results}a-d shows results for negative the log likelihood (NLL) of the convolutional model on  MNIST with masking rate $29\%$ and 60,000 training samples. Figure \ref{fig:main:cifar_cnn_nll_tradeoff_results}e-h shows results for NLL of the convolutional model on CIFAR10 with 20,000 samples. We select these settings as illustrative of more difficult cases for posterior predictive distribution distillation. We plot NLL vs FLOPS and NLL vs storage for all points encountered in each search. The solid blue line indicates the Pareto frontier. 

Figure \ref{fig:main:cnn_nll_tradeoff_results} shows results for the negative log likelihood (NLL) of the convolutional model on  MNIST with masking rate $29\%$ and 60,000 training samples. We select this setting as illustrative of a difficult case for posterior predictive distribution distillation. We plot NLL vs FLOPS and NLL vs storage for all points encountered in each search. The solid blue line indicates the Pareto frontier. 

% First, we note that the baseline student model (with architecture matching the teacher) from Experiment 2 on MNIST achieves an NLL of $0.469$ at approximately $0.48 \times 10^6$ FLOPs and $0.03 \times 10^6$ parameters on this configuration of the data set, while for CIFAR10, the baseline student model achieves an NLL of $1.2$ at approximately $2.5 \times 10^6$ FLOPs and $0.03 \times 10^6$ parameters. We can see that both methods for selecting student architectures provide a highly significant improvement over the baseline student architectures. On MNIST, the NLL is reduced to $0.30$ while for CIFAR10 it is reduced to $0.90$.
% Further, we can also see that the group $\ell_1/\ell_2$ approach is able to obtain much better NLL at the same computation and storage cost relative to the exhaustive search method. Lastly, the group $\ell_1/\ell_2$ method is able to obtain models on MNIST at less than $50\%$ the computational cost needed by the baseline model with only a small loss in performance. On CIFAR10 this is not the case, but we expect that further refinement of the range of regularization parameter strengths used will expose a greater space of trade-offs. Results for other models and distillation targets show similar trends and are presented in Appendix \ref{appendix:experiments}. Additional experimental details are given in Appendix \ref{appendix:models}.

First, we note that the baseline student model (with architecture matching the teacher) from Experiment 2 on MNIST achieves an NLL of $0.469$ at approximately $0.48 \times 10^6$ FLOPs and $0.03 \times 10^6$ parameters on this configuration of the data set. We can see that both methods for selecting student architectures provide a highly significant improvement over the baseline student architectures. On MNIST, the NLL is reduced to $0.30$.
Further, we can also see that the group $\ell_1/\ell_2$ approach is able to obtain much better NLL at the same computation and storage cost relative to the exhaustive search method. Lastly, the group $\ell_1/\ell_2$ method is able to obtain models on MNIST at less than $50\%$ the computational cost needed by the baseline model with only a small loss in performance. Results for other models and distillation targets show similar trends and are presented in Appendix \ref{appendix:experiments}. Additional experimental details are given in Appendix \ref{appendix:models}.

In summary, the key finding of this experiment is that the capacity of the student model significantly impacts distillation performance, and student model architecture optimization methods are needed to achieve a desired speed-storage-accuracy trade-off. 

\begin{table}[htbp]
\centering
\caption{In-distribution Test set metrics comparison using $U_s$ and largest student model obtained using width multiplier.}
\label{tab:test-metrics-comparison-us-large-model}
\resizebox{\linewidth}{!}{
\begin{tabular}{cccccc}
\toprule
\begin{tabular}[c]{@{}c@{}}Model/\\ Dataset\end{tabular} &
  \begin{tabular}[c]{@{}c@{}}NLL\\ (Ensemble)\end{tabular} &
  \begin{tabular}[c]{@{}c@{}}NLL\\ (GPED)\end{tabular} &
  \begin{tabular}[c]{@{}c@{}}NLL\\ (EnD\textsuperscript{2})\end{tabular} &
  \begin{tabular}[c]{@{}c@{}}MAE \\ Entropy\\ (GPED)\end{tabular} &
  \begin{tabular}[c]{@{}c@{}}MAE \\Entropy\\ (EnD\textsuperscript{2})\end{tabular} \\ \midrule
\begin{tabular}[c]{@{}c@{}}FCNN/\\ MNIST\end{tabular}  & 0.362 & 0.408 & 0.415 & 0.069 & 0.105 \\ \midrule
\begin{tabular}[c]{@{}c@{}}CNN/\\ MNIST\end{tabular}   & 0.269 & 0.296 & 0.321 & 0.086 & 0.106 \\ \midrule
\begin{tabular}[c]{@{}c@{}}CNN/\\ CIFAR10\end{tabular} & 0.799 & 0.859 & 0.907 & 0.146 & 0.328 \\ \bottomrule
\end{tabular}
}
\end{table} 

\textbf{Experiment 4: Uncertainty Quantification for Downstream Tasks} 
As noted earlier, uncertainty quantification and decomposition is an important application of Bayesian posterior predictive inference. In this set of experiments, we evaluate our method on two downstream applications: out-of-distribution detection and uncertainty-based ranking. We compare the GPED framework to the full Monte Carlo ensemble as well as to an adaptation of Ensemble Distribution Distillation (EnD\textsuperscript{2}) \citep{malinin2020ensemble}. In particular,  \citet{malinin2020ensemble} materialize a complete ensemble, which is not feasible in our case due to the large number of models in the Bayesian ensemble ($\sim 10^5$ models). We instead use Algorithm \ref{algorithm:ped} with the Dirichlet log likelihood distillation loss used by \citet{malinin2020ensemble} (see Appendix A.3 for EnD\textsuperscript{2} implementation details). Additionally, we modify our student models to distill both the predictive distribution and expected data uncertainty in a single model. 

Before assessing the performance of these methods on downstream tasks, we first compare their performance in terms of negative log likelihood and MAE on the posterior predictive distribution and expected data uncertainty distillation tasks. We use the same dataset augmentation as in the previous experiment. We compare the GPED and EnD\textsuperscript{2} methods using the $U_o$ and $U_s$ as well as for small and large model sizes. Note that for distilling entropy under our method in this section, we always use $U_o$ estimator. Wherever the $U_s$ estimator is mentioned for our method in this section of experiments, it is only applied to distilling predictive means. In Table \ref{tab:test-metrics-comparison-us-large-model} we compare different distillation methods for different model-dataset combinations. These results correspond to the $U_s$ estimator and the largest student model. As an illustration, we present joint and marginal expected data uncertainty distribution plots in Figure \ref{fig:joint_distribution_plots} that correspond to the results in Table \ref{tab:test-metrics-comparison-us-large-model}. These figures show how GPED and EnD\textsuperscript{2} compare against the Bayesian ensemble on a data case-by-data case basis.
Additional results are presented in Tables [\ref{tab:test-metrics-comparison-us-small-model}- \ref{tab:test-metrics-comparison-uo-large-model}] and Figure \ref{fig:gped_pn_comparison}.  The key result of these experiments is that the GPED framework consistently performs better than EnD\textsuperscript{2} across all metrics on the test datasets. 

\textbf{Out-of-distribution detection: } OOD detection has garnered a lot interest in the deep learning community as it is as a practical challenge during deployment of deep models. In this experiment, we use the measures of total uncertainty and knowledge uncertainty for detecting OOD inputs.
\begin{table}[]
\centering
\caption{AUROC for OOD Detection using $U_s$ and largest student model obtained using width multiplier.}
\label{tab:ood-auroc-us-large-model}
\resizebox{\linewidth}{!}{
\begin{tabular}{ccccc}
\toprule
\begin{tabular}[c]{@{}c@{}}Model \& Train Data/\\ OOD Data\end{tabular} & Uncertainty & Ensemble & \begin{tabular}[c]{@{}c@{}}GPED\\ (ours)\end{tabular} & EnD\textsuperscript{2} \\ \midrule
\multirow{2}{*}{\begin{tabular}[c]{@{}c@{}}FCNN-MNIST/\\ KMNIST\end{tabular}} & Total     & 0.929 & 0.867 & 0.816 \\ \cline{2-5} 
                                                                              & Knowledge & 0.976 & 0.928 & 0.899 \\ \midrule
\multirow{2}{*}{\begin{tabular}[c]{@{}c@{}}FCNN-MNIST/\\ notMNIST\end{tabular}} & Total     & 0.944 & 0.670 & 0.652 \\ \cline{2-5} 
                                                                              & Knowledge & 0.990 & 0.762 & 0.681 \\ \midrule

\multirow{2}{*}{\begin{tabular}[c]{@{}c@{}}CNN-MNIST/\\ KMNIST\end{tabular}} & Total     & 0.894 & 0.882 & 0.881 \\ \cline{2-5} 
                                                                              & Knowledge & 0.956 & 0.932 & 0.952 \\ \midrule

\multirow{2}{*}{\begin{tabular}[c]{@{}c@{}}CNN-MNIST/\\ notMNIST\end{tabular}} & Total     & 0.888 & 0.882 & 0.860 \\ \cline{2-5} 
                                                                              & Knowledge & 0.946 & 0.934 & 0.939 \\ \midrule

\multirow{2}{*}{\begin{tabular}[c]{@{}c@{}}CNN-CIFAR10/\\ TIM\end{tabular}}      & Total     & 0.729 & 0.762 & 0.721 \\ \cline{2-5} 
                                                                              & Knowledge & 0.796 & 0.808 & 0.792 \\ \midrule
                                                                              
\multirow{2}{*}{\begin{tabular}[c]{@{}c@{}}CNN-CIFAR10/\\ LSUN\end{tabular}}      & Total     & 0.790 & 0.779 & 0.747 \\ \cline{2-5} 
                                                                              & Knowledge & 0.752 & 0.767 & 0.713 \\ \bottomrule
\end{tabular}
}
\end{table}
OOD detection is a binary classification problem where we utilize a measure of uncertainty to classify an input as in-distribution or out-of-distribution based on a threshold. For our experiments, we use four OOD datasets: KMNIST \citep{clanuwat2018deep}, notMNIST \citep{Yaroslav2011notmnist}, TinyImageNet (TIM) (CS231N, 2017), and SVHN \citep{netzer2011reading}. Additional experimental details are given in the Appendix A.4. We run our experiments for different combinations of models, in-distribution datasets, out-of-distribution datasets, model architectures, and estimators used for distilling the predictive distribution under the proposed framework as well as for the EnD\textsuperscript{2} framework. We report example OOD detection results using the $U_s$ estimator and the largest student model in Table \ref{tab:ood-auroc-us-large-model}. Our overall results show that GPED outperforms EnD\textsuperscript{2} in 75\% of cases across all experimental settings considered (additional results are given in Tables [\ref{tab:ood-auroc-us-small-model}-\ref{tab:ood-auroc-uo-large-model}] in Appendix B). 

% Our first observation is that our GPED framework consistently gives superior performance as compared to the EnD\textsuperscript{2} framework across all model and dataset combinations on this task. We also see observe that our GPED framework yields better performance than ensemble for the MNIST/fMNIST case. Thus, a key takeaway from this experiment is that our framework shows promising results on OOD detection while operating under significantly low storage-computation regime as compared to the ensemble. 
\textbf{Uncertainty-Based Ranking:} Another important application of Bayesian neural networks is ranking instances based on uncertainty. Such rankings are used in active learning and other human-in-the-loop decision systems to prioritize uncertain instances for labeling or analysis by human decision makers. This task is sensitive to the correct rank order of in-distribution instances by uncertainty level, where as the OOD task is only sensitive to the existence of a threshold that separates in and out of distribution instances. To assess how well our distillation framework preserves the relative ranking between the inputs when compared to the full Bayesian ensemble, we compute the Normalized Discounted Cumulative  Gain (nDCG) score \citep{Jrvelin2002CumulatedGE} for total uncertainty and knowledge uncertainty. A higher nDCG score implies that the correct ranking of inputs is better preserved under the distillation framework. For our experiments, we asses nDCG@20. In Table \ref{tab:ndcg-comparison-us-large-model},  we report the nDCG scores using the $U_s$ estimator and largest student model as example results. Overall, GPED outperforms  EnD\textsuperscript{2} in 91\% of settings considered (additional ranking results are given in Tables [\ref{tab:ndcg-comparison-us-small-model}-\ref{tab:ndcg-comparison-u0-large-model}] in Appendix B).

\begin{table}[]
\centering
\caption{nDCG@20 out of 100 randomly selected test inputs using $U_s$ estimator and largest student model . Results reported as mean $\pm$ std. dev. over 500 trials.}
\label{tab:ndcg-comparison-us-large-model}
\resizebox{\linewidth}{!}{
\begin{tabular}{cccc}
\toprule
Model \& Data                        & Uncertainty & \begin{tabular}[c]{@{}c@{}}GPED\\ (ours)\end{tabular} & EnD\textsuperscript{2}         \\ \midrule
\multirow{2}{*}{FCNN-MNIST} & Total       & 0.954 $\pm$ 0.02                                         & 0.946 $\pm$ 0.021 \\ \cline{2-4} 
                            & Knowledge   & 0.924 $\pm$ 0.03                                          & 0.941 $\pm$ 0.028 \\ \midrule

\multirow{2}{*}{CNN-MNIST} & Total       & 0.929 $\pm$ 0.034                                          & 0.916 $\pm$ 0.032 \\ \cline{2-4} 
                            & Knowledge   & 0.888 $\pm$ 0.032                                          & 0.876 $\pm$ 0.045 \\ \midrule

\multirow{2}{*}{CIFAR10}   & Total       & 0.935 $\pm$ 0.022                                          & 0.919 $\pm$ 0.027 \\ \cline{2-4} 
                            & Knowledge   & 0.885 $\pm$ 0.033                                         & 0.889 $\pm$ 0.034 \\ \bottomrule
\end{tabular}
}
\end{table}

% \subsection{Multi-Output Student Models}
% In all the previous experiments, the student model was trained to distill a single posterior expectation quantity. Thus, in cases where we want to obtain estimates of multiple posterior expectations, we need to store as many models and compute as many forward passes through the student as the number of posterior expectations desired. Similar to the previous set of experiments, we produce the performance tradeoff against storage and computation cost for different configurations of models, datasets and tasks.

%\input{experiment_masked_MNIST.tex}

%\input{improving_distillation.tex}

\section{Conclusions \& Future Directions}
We have presented a  framework for distilling expectations with respect to the Bayesian posterior distribution of a deep neural network that significantly generalizes the Bayesian Dark Knowledge approach. Our results show that posterior distillation performance can be highly sensitive to the architecture of the student model, but that  architecture search methods can identify student model architectures with improved speed-storage-accuracy trade-offs. We have also demonstrated that the proposed approach performs well on downstream tasks that leverage entropy distillation for uncertainty decomposition.
There are many directions for future work including considering  the distillation of a broader class of posterior statistics, developing more advanced architecture search methods, and applying the framework to larger  models.
\section*{Acknowledgments}
This work was partially supported by the US Army Research Laboratory under cooperative agreement W911NF-17-2-0196. The   views   and   conclusions  contained  in  this  document  are  those  of  the  authors  and  should  not  be interpreted as representing the official policies,  either  expressed  or  implied,  of  the  Army  Research  Laboratory  or  the  US  government.

\bibliography{references_short}
\bibliographystyle{abbrvnat}

\appendix
\section{Datasets and Model Details}
\subsection{Datasets}
\label{appendix:datasets}
As noted earlier in the paper, the original empirical investigation of Bayesian Dark Knowledge for classification focused on the MNIST data set \citep{lecun1998mnist}. However, the models fit to the MNIST data set have very low posterior uncertainty and we argue that it is thus a poor benchmark for assessing the performance of posterior distillation methods. In this section, we investigate two orthogonal modifications of the standard MNIST data set to increase uncertainty: reducing the training set size and masking regions of the input images. Our goal is to produce a range of benchmark problems with varying posterior predictive uncertainty. We also use the CIFAR10 data  set \citep{krizhevsky2009learning} in our experiments and employ the same subsampling technique.

\textbf{MNIST:} The full MNIST dataset consists of 60,000 training images and 10,000 test images, each of size $28 \times 28$, distributed among 10 classes \citet{lecun1998mnist}. As a first manipulation, we consider sub-sampling the labeled training data to include 10,000, 20,000, 30,000 or all 60,000 data cases in the primary data set  $\pazocal{D}$ when performing posterior sampling for the teacher model. Importantly, we use all 60,000 unlabeled training cases in the distillation data set  $\pazocal{D}'$. This allows us de-couple the impact of reduced labeled training data on posterior predictive distributions from the effect of the amount of unlabeled data available for distillation. 

\begin{figure}[htbp]
    \centering
    \subfigure[Original images]{\includegraphics[width=0.2\textwidth]{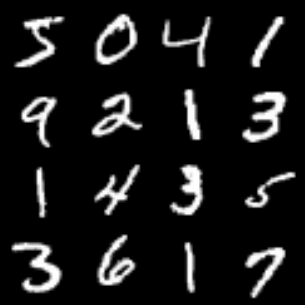}}
    \hspace{1cm}
    \subfigure[Processed images]{\includegraphics[width=0.2\textwidth]{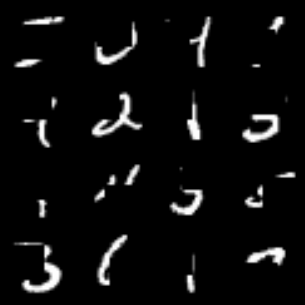}}
    \caption{Example  MNIST data after masking with $m=14$.\vspace{-1em}}
    \label{fig:mnist}
\end{figure}

As a second manipulation, we generate images with occlusions by randomly masking out parts of each available training and test image. For generating such images, we randomly choose a square $m\times m$ region (mask) and set the value for pixels in that region to 0. Thus, the masking rate for a $28 \times 28$ MNIST image corresponding to the mask of size $m\times m$ is given by $r = \frac{m \times m}{28 \times 28}$. We illustrate original and masked data in Figure \ref{fig:mnist}. We consider a range of square masks resulting in masking rates between 0\% and 86.2\%.

\textbf{CIFAR10:} The full CIFAR10 dataset consists of 50,000 training images and 10,000 test images, each of size $32 \times 32$ pixels. We sub-sample the data into a primary training sets  $\pazocal{D}$ containing 10,000, 20,000, and 50,000 images. As with MNIST, the sub-sampling is limited to training the teacher model only and we utilize all the 50,000 unlabeled training images in the distillation  data set  $\pazocal{D}'$.

\subsection{Models}
\label{appendix:models}
To demonstrate the generalizability of our methods to a range of model architectures, we run our experiments with both fully-connected, and convolutional neural networks. We note that our goal in this work is not to evaluate the GPED framework on state-of-the-art architectures, but rather to provide illustrative results and establish methodology for assessing the impact of several factors including the level of uncertainty and the architecture of the student model.

\textbf{Teacher Models:} We begin by defining the architectures used for the teacher model as follows:
\begin{enumerate}
    \item \textbf{FCNN (MNIST)}: We use a 3-layer fully connected neural network. The architecture used is: Input(784)-FC(400)-FC(400)-FC(output). This matches the architecture used by \cite{balan2015bayesian}.
    \item \textbf{CNN (MNIST)}: For a CNN, we use two consecutive sets of 2D convolution and max-pooling layers, followed by two fully-connected layers. The architecture used is: Input(1, (28,28))-Conv(num\_kernels=10, kernel\_size=4, stride=1) - MaxPool(kernel\_size=2) - Conv(num\_kernels=20, kernel\_size=4, stride=1) - MaxPool(kernel\_size=2) - FC (80) - FC (output).
    \item \textbf{CNN (CIFAR10)}: Similar to the CNN architecture used for MNIST, we use two consecutive sets of 2D convolution and max-pooling layers followed by fully-connected layers.  Conv(num\_kernels=16, kernel\_size=5) - MaxPool(kernel\_size=2) - Conv(num\_kernels=32, kernel\_size=5) - MaxPool(kernel\_size=2) - FC(200) - FC (50) - FC (output).
\end{enumerate}
In the architectures mentioned above, the ``output" size will change depending on the expectation that we're distilling. For classification, the output size will be $10$ for both datasets, while for the case of entropy, it will be $1$. We use \emph{ReLU} non-linearities everywhere between the hidden layers. For the final output layer, \emph{softmax} is used for classification. In the case of entropy, we use an exponential activiation to ensure positivity. 

\textbf{Student Models:} The student models used in our experiments use the above mentioned architectures as the base architecture. For explicitly searching the space of the student models, we use a set of width multipliers starting from the teacher architecture. The space of student architectures corresponding to each teacher model defined earlier is given below. The width multiplier values of $K_1$ and $K_2$ are determined differently for each of the experiments, and thus will be mentioned in later sections. 
\begin{enumerate}
    \item \textbf{FCNN (MNIST)}: Input(784)-FC($400\cdot K_1$)-FC($400\cdot K_2$)-FC(output).
    \item \textbf{CNN (MNIST)}: Input(1, (28,28))-Conv(num\_kernels=$\floor{10\cdot K_1}$, kernel\_size=4, stride=1) - MaxPool(kernel\_size=2) - Conv(num\_kernels=$\floor{20\cdot K_1}$, kernel\_size=4, stride=1) - MaxPool(kernel\_size=2) - FC ($\floor{80\cdot K_2}$) - FC (output).
    \item \textbf{CNN (CIFAR10)}: Input(3, (32,32))-Conv(num\_kernels=$\floor{16\cdot K_1}$, kernel\_size=5) - MaxPool(kernel\_size=2) - Conv(num\_kernels=$\floor{16\cdot K_1}$, kernel\_size=5) - MaxPool(kernel\_size=2) - FC ($\floor{200\cdot K_2}$) - FC ($\floor{50\cdot K_2}$) - FC (output).
\end{enumerate}

\textbf{Model and Distillation Hyper-Parameters}: We run the distillation procedure using the following hyperparameters: fixed teacher learning rate $\eta_t = 4 \times 10^{-6}$ for models on MNIST and $\eta_t = 2 \times 10^{-6}$ for models on CIFAR10, teacher prior precision $\tau = 10$, initial student learning rate $\alpha_s = 10^{-3}$, student dropout rate $p=0.5$ for fully-connected models on MNIST (and zero otherwise), burn-in iterations $B=1000$ for MNIST and $B=10000$ for CIFAR10, thinning interval $H = 100$ for distilling predictive means and $H=10$ for distilling  entropy values, and total training iterations $T= 10^6$. For training the student model, we use the \emph{Adam} algorithm (instead of plain steepest descent as indicated in Algorithm 1) and set a learning schedule for the student such that it halves its learning rate every 200 epochs for models on MNIST, and every 400 epochs for models on CIFAR10. Also, note that we only apply the regularization function $R(\phi_s)$ while doing Group $\ell_1/\ell_2$ pruning. Otherwise, we use dropout as indicated before.

\begin{figure*}[t]
    \centering
    \subfigure[]{\includegraphics[width=0.32\textwidth]{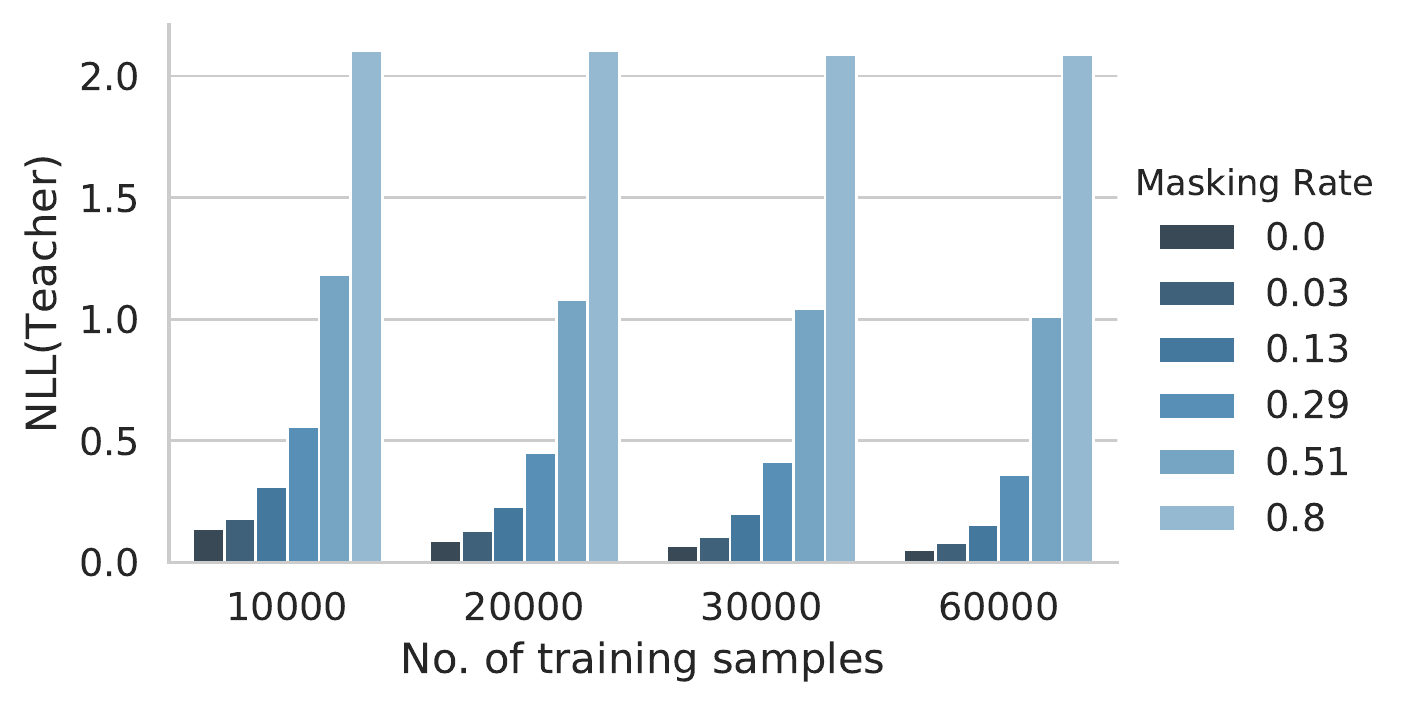}}
    \subfigure[]{\includegraphics[width=0.32\textwidth]{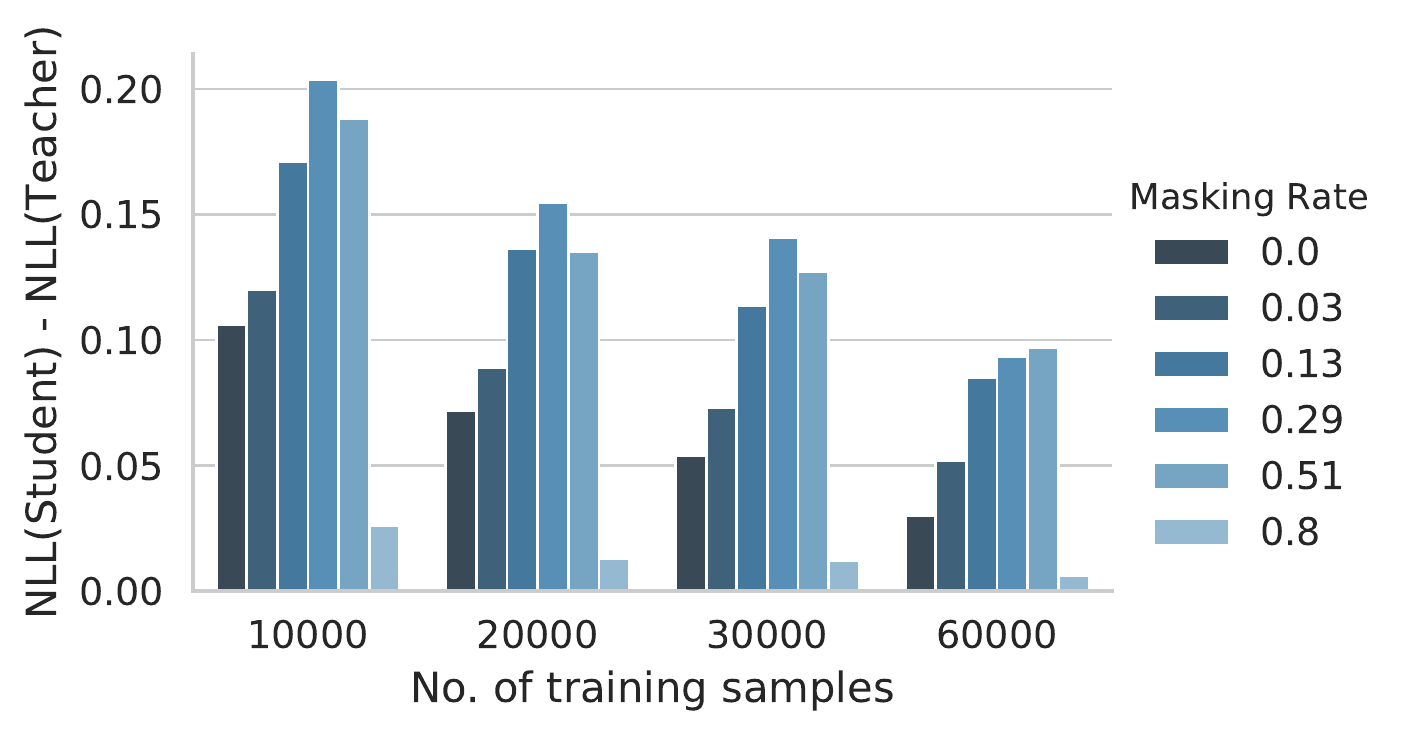}}
    \subfigure[]{\includegraphics[width=0.32\textwidth]{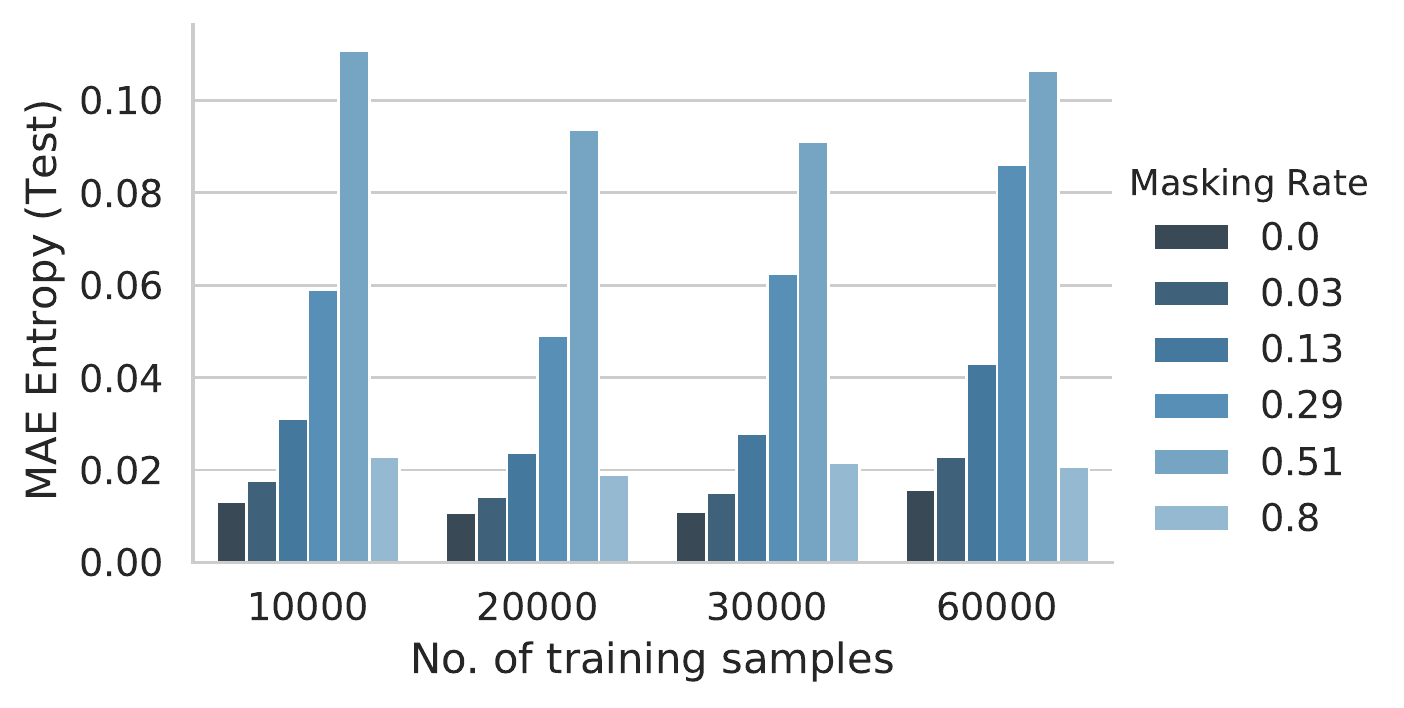}}
    \caption{Distillation performance using Fully-Connected Networks on MNIST while varying data set size and masking rate. (a) Test negative log likelihood of the teacher posterior predictive distribution. (b) Difference in test negative log likelihood between teacher and student posterior predictive distribution estimates. (c) Difference between teacher and student posterior entropy estimates on test data set.}
    \label{fig:fcnn_mnist_robustness_results}
\end{figure*} 

\begin{figure*}[t]
%
%
%\begin{figure}[htbp]
%    \centering
    \subfigure[]{\includegraphics[width=0.33\textwidth]{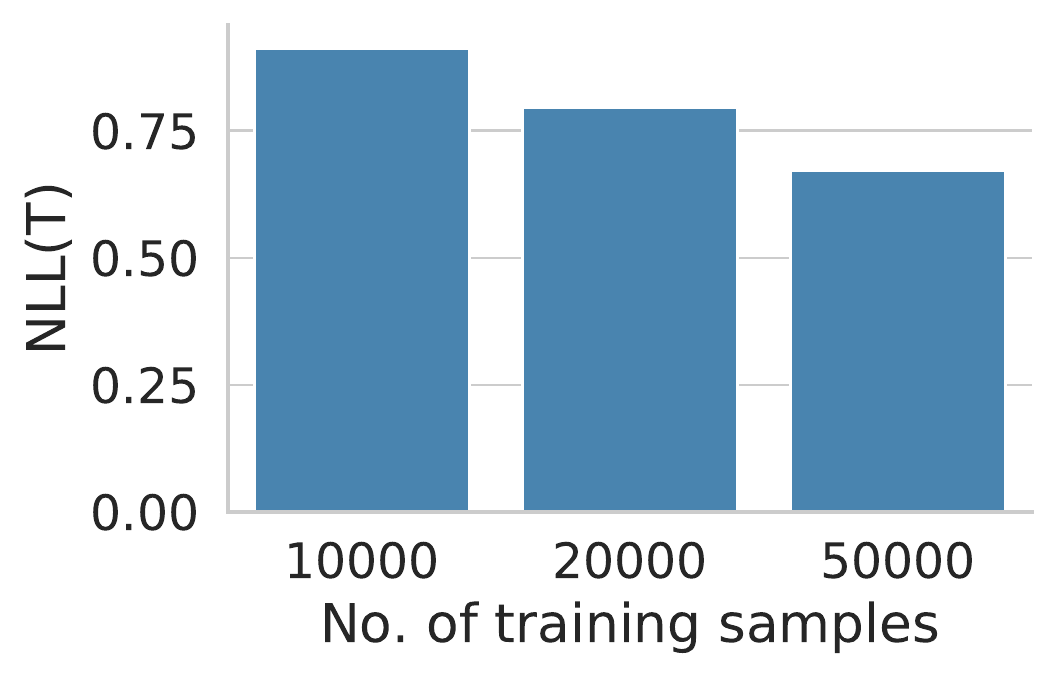}}
    \subfigure[]{\includegraphics[width=0.33\textwidth]{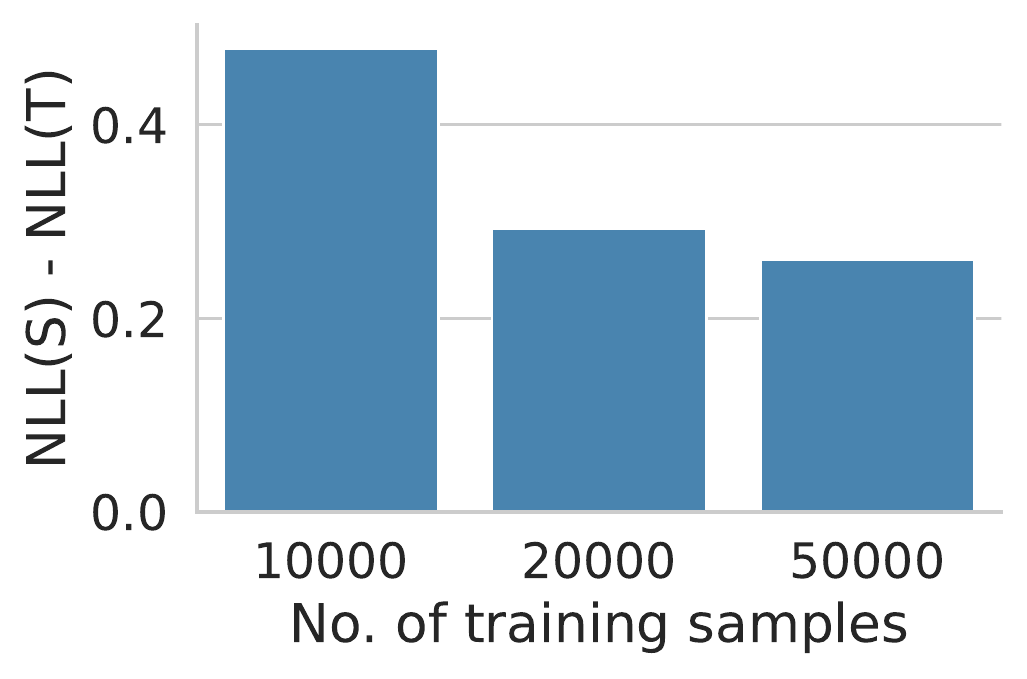}
    }
    \subfigure[]{\includegraphics[width=0.32\textwidth]{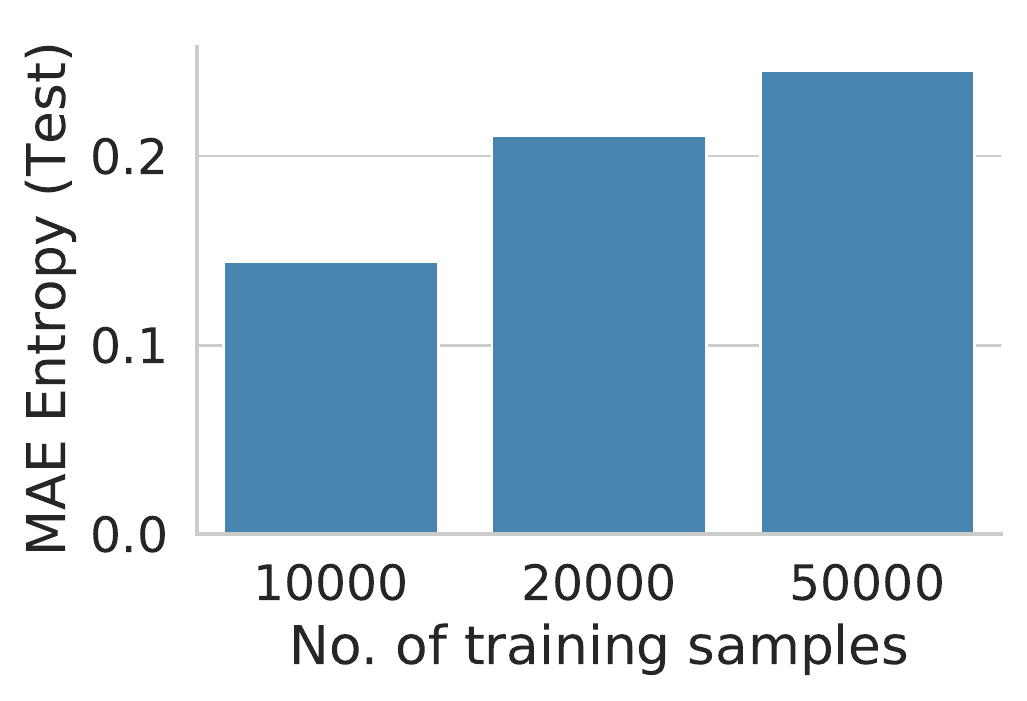}}
    \caption{Distillation performance using CNNs on CIFAR10 while varying data set size. (d) Test negative log likelihood of the teacher posterior predictive distribution. (e) Difference in test negative log likelihood between student and teacher posterior predictive distribution estimates. (f) Difference between teacher and student posterior entropy estimates on test data set. In the plots above, S denotes the student and T denotes the teacher.}
    \label{fig:cnn_cifar_robustness_results}

\end{figure*} 

\textbf{Hyper-parameters for Group $\ell_1/\ell_2$ pruning experiments}: For experiments involving group $\ell_1/\ell_2$ regularizer, the regularization strength values $\lambda$ are chosen from a log-scale ranging from $10^{-8}$ to $10{-3}$. When using Group $\ell_1/\ell_2$ regularizer, we do not use dropout for the student model. The number of fine-tuning epochs for models on MNIST and CIFAR100 are 600 and 800 respectively. At the start of fine-tuning, we also reinitialize the student learning rate $\alpha_t = 10^{-4}$ for fully-connected models and $\alpha_t = 10^{-3}$ for convolutional models. The magnitude threshold for pruning is $\epsilon=10^{-3}$.

\subsection{Ensemble Distribution Distillation (EnD\textsuperscript{2}) \citep{malinin2020ensemble}}

In this framework, distillation of ensembles is performed into a prior network (PN), which predicts the concentration parameters of a dirichlet distribution as output given an input data point. The prior network is learned by maximizing the log likelihood given the predictive distribution of the teacher ensemble. 
\begin{align}
\label{eq:pn_mle}
\pazocal{L}\left(\phi, \pazocal{D'}\right) &=-\mathbb{E}_{p(\mathbf{x})}\left[\mathbb{E}_{p(y | \mathbf{x})}[\ln \mathrm{p}_{dir}(\boldsymbol{\pi} | \boldsymbol{x} ; \phi)]\right] \notag \\ 
&=-\frac{1}{N} \sum_{i=1}^{N}\Biggl[\ln \Gamma\left(\hat{\alpha}_{0}^{(i)}\right)-\sum_{c=1}^{K} \ln \Gamma\left(\hat{\alpha}_{c}^{(i)}\right) \notag \\
&+ \frac{1}{M} \sum_{m=1}^{M} \sum_{c=1}^{K}\left(\hat{\alpha}_{c}^{(i)}-1\right) \ln( p(y=c | \mathbf{x}, \theta_m) ) \Biggr] 
\end{align}
Here, $\{\theta_m\}_{m=1}^M$ denotes the teacher ensemble, $\boldsymbol{\pi}$ is the categorial distribution obtained by one of the teacher models (i.e. $\boldsymbol{\pi} = p(y | \mathbf{x}, \theta_m)$). It is evident that to estimate the expectation mentioned above, we can use both $U_o$ and $U_s$ estimators. As we noted earlier, materializing all the samples is not feasible in our case due to the size of the resulting Bayesian ensemble ($\sim 10^5$ models). To learn prior network without materializing all the samples drawn using SGLD, we again use Algorithm \ref{algorithm:ped}, but replace the loss function on line 12 with the Equation \ref{eq:pn_mle}. 

\subsection{Additional Details on Experiment 4: Uncertainty Quantification for Downstream tasks} 

We use the same dataset augmentations as used earlier Experiment 3. For assessing the performance on downstream tasks wrt student model architecture, we use a base student model with the same architecture as the teacher and the largest width multiplier explored in Experiment 3. As noted earlier, we augment the student models under our proposed framework to distill both predictive distribution as well as expected entropy (thus making our network's output dimensionality $C+1$, where $C$ denotes the number of classes). The output dimensionality of prior network is $C$ (for $C$ different concentration parameters). We use exponential activation at expected entropy output as well as prior network output to ensure the positive constraint. We additionally use a temperature value of $\tau_s=2.5$ while training all the prior network based models. \citet{malinin2020ensemble} suggest training the prior networks with a temperature annealing schedule, however we find that in our experiments prior networks achieve better performance in terms of log likelihood while using a fixed temperature. Dropout rate for CNN models is set at $p=0.3$, while FCNN models have a dropout rate of $p=0.5$. The rest of the experimental details remain the same as stated in Experiment 3.

\section{Supplemental Experiments and Results}
\label{appendix:experiments}

\subsection{Supplemental Results for Experiment 2: Robustness to Uncertainty} 
% \begin{figure}[htbp]
%     \centering
%     \subfigure[]{\includegraphics[width=0.31\textwidth]{figures/mnist_cnn/cnn_masking_teacher_nll_results.pdf}}
%     \subfigure[]{\includegraphics[width=0.31\textwidth]{figures/mnist_cnn/cnn_masking_delta_nll_results.pdf}}
%     \subfigure[]{\includegraphics[width=0.31\textwidth]{figures/mnist_cnn/cnn_masking_delta_entropy_results.pdf}}
%     \caption{Distillation performance while using CNNs on augmented MNIST}
%     \label{fig:cnn_mnist_robustness_results}
% \end{figure} 
In Figure \ref{fig:fcnn_mnist_robustness_results}, we demonstrate the results of Experiment 2 (Section 4.3), on fully-connected networks for MNIST. Additionally, in Tables [\ref{tab:estimator_comparison_mnist_cnn}-\ref{tab:estimator_comparison_cifar_cnn}], we provide a performance comparison between $U_o$ and $U_s$ estimators while distilling posterior expectations for all model-data set combinations. We follow the same experimental configurations as in Experiment 2.

\begin{table*}[t]
\centering
\caption{Performance comparison between $U_o$ and $U_s$ estimators for convolutional neural network on MNIST. The NLL results correspond to the case of distilling the posterior predictive distribution while the MAE on entropy results correspond to the case of distilling the expectation of predictive entropy.}
\begin{tabular}{ccccccc}
\toprule
\begin{tabular}[c]{@{}c@{}}Num. training \\ samples\end{tabular} & \begin{tabular}[c]{@{}c@{}}Masking \\ rate\end{tabular} & \begin{tabular}[c]{@{}c@{}}NLL\\ (Teacher)\end{tabular} & \begin{tabular}[c]{@{}c@{}}NLL\\ (Student, $U_o$)\end{tabular} & \begin{tabular}[c]{@{}c@{}}NLL\\ (Student, $U_s$)\end{tabular} & \begin{tabular}[c]{@{}c@{}}MAE\\ (Entropy, $U_o$)\end{tabular} & \begin{tabular}[c]{@{}c@{}}MAE\\ (Entropy, $U_s$)\end{tabular} \\
\midrule
\multirow{7}{*}{10000} & 0 & 0.048 & 0.214 & 0.218 & 0.025 & 0.030 \\
 & 0.03 & 0.069 & 0.274 & 0.274 & 0.033 & 0.038 \\
 & 0.13 & 0.161 & 0.509 & 0.509 & 0.058 & 0.069 \\
 & 0.29 & 0.394 & 0.902 & 0.907 & 0.115 & 0.129 \\
 & 0.51 & 1.099 & 1.615 & 1.630 & 0.194 & 0.170 \\
 & 0.8 & 2.298 & 2.301 & 2.301 & 0.016 & 0.019 \\
 \midrule
 & 0 & 0.034 & 0.126 & 0.126 & 0.020 & 0.021 \\
\multirow{5}{*}{20000} & 0.03 & 0.054 & 0.180 & 0.181 & 0.026 & 0.030 \\
 & 0.13 & 0.123 & 0.342 & 0.344 & 0.053 & 0.066 \\
 & 0.29 & 0.326 & 0.684 & 0.697 & 0.104 & 0.122 \\
 & 0.51 & 1.050 & 1.369 & 1.378 & 0.145 & 0.150 \\
 & 0.8 & 2.298 & 2.300 & 2.299 & 0.016 & 0.020 \\
 \midrule
\multirow{6}{*}{30000} & 0 & 0.028 & 0.084 & 0.086 & 0.017 & 0.019 \\
 & 0.03 & 0.044 & 0.132 & 0.134 & 0.024 & 0.027 \\
 & 0.13 & 0.106 & 0.292 & 0.294 & 0.051 & 0.061 \\
 & 0.29 & 0.300 & 0.620 & 0.618 & 0.106 & 0.120 \\
 & 0.51 & 1.044 & 1.307 & 1.308 & 0.130 & 0.141 \\
 & 0.8 & 2.296 & 2.297 & 2.296 & 0.017 & 0.021 \\
 \midrule
\multirow{6}{*}{60000} & 0 & 0.022 & 0.053 & 0.053 & 0.016 & 0.017 \\
 & 0.03 & 0.035 & 0.088 & 0.090 & 0.025 & 0.026 \\
 & 0.13 & 0.090 & 0.219 & 0.221 & 0.049 & 0.058 \\
 & 0.29 & 0.267 & 0.463 & 0.472 & 0.108 & 0.120 \\
 & 0.51 & 1.024 & 1.184 & 1.187 & 0.118 & 0.127 \\
 & 0.8 & 2.297 & 2.297 & 2.297 & 0.020 & 0.023  \\
 \bottomrule
\end{tabular}
\label{tab:estimator_comparison_mnist_cnn}
\end{table*}

% Please add the following required packages to your document preamble:
% \usepackage{multirow}
\begin{table*}[]
\centering
\caption{Performance comparison between $U_o$ and $U_s$ estimators for fully-connected network on MNIST. The NLL results correspond to the case of distilling the posterior predictive distribution while the MAE on entropy results correspond to the case of distilling the expectation of predictive entropy.}
\begin{tabular}{ccccccc}
\toprule
\begin{tabular}[c]{@{}c@{}}Num. training \\ samples\end{tabular} & \begin{tabular}[c]{@{}c@{}}Masking \\ rate\end{tabular} & \begin{tabular}[c]{@{}c@{}}NLL\\ (Teacher)\end{tabular} & \begin{tabular}[c]{@{}c@{}}NLL\\ (Student, $U_o$)\end{tabular} & \begin{tabular}[c]{@{}c@{}}NLL\\ (Student, $U_s$)\end{tabular} & \begin{tabular}[c]{@{}c@{}}MAE\\ (Entropy, $U_o$)\end{tabular} & \begin{tabular}[c]{@{}c@{}}MAE\\ (Entropy, $U_s$)\end{tabular} \\
\midrule
\multirow{6}{*}{10000} & 0 & 0.137 & 0.184 & 0.243 & 0.013 & 0.018 \\
 & 0.03 & 0.180 & 0.233 & 0.300 & 0.018 & 0.023 \\
 & 0.13 & 0.312 & 0.389 & 0.483 & 0.031 & 0.040 \\
 & 0.29 & 0.556 & 0.637 & 0.760 & 0.059 & 0.089 \\
 & 0.51 & 1.183 & 1.229 & 1.371 & 0.111 & 0.135 \\
 & 0.8 & 2.103 & 2.111 & 2.129 & 0.023 & 0.019 \\
 \midrule
\multirow{6}{*}{20000} & 0 & 0.089 & 0.115 & 0.161 & 0.011 & 0.015 \\
 & 0.03 & 0.131 & 0.165 & 0.220 & 0.014 & 0.021 \\
 & 0.13 & 0.230 & 0.280 & 0.366 & 0.024 & 0.042 \\
 & 0.29 & 0.452 & 0.510 & 0.607 & 0.049 & 0.104 \\
 & 0.51 & 1.080 & 1.120 & 1.215 & 0.094 & 0.112 \\
 & 0.8 & 2.104 & 2.108 & 2.117 & 0.019 & 0.021 \\
 \midrule
\multirow{6}{*}{30000} & 0 & 0.071 & 0.083 & 0.124 & 0.011 & 0.014 \\
 & 0.03 & 0.107 & 0.129 & 0.180 & 0.015 & 0.021 \\
 & 0.13 & 0.201 & 0.243 & 0.314 & 0.028 & 0.052 \\
 & 0.29 & 0.414 & 0.459 & 0.555 & 0.062 & 0.105 \\
 & 0.51 & 1.044 & 1.082 & 1.172 & 0.091 & 0.105 \\
 & 0.8 & 2.089 & 2.092 & 2.101 & 0.022 & 0.023 \\
 \midrule
\multirow{6}{*}{60000} & 0 & 0.052 & 0.054 & 0.082 & 0.016 & 0.020 \\
 & 0.03 & 0.081 & 0.094 & 0.133 & 0.023 & 0.034 \\
 & 0.13 & 0.155 & 0.186 & 0.240 & 0.043 & 0.068 \\
 & 0.29 & 0.360 & 0.398 & 0.471 & 0.086 & 0.109 \\
 & 0.51 & 1.010 & 1.033 & 1.107 & 0.106 & 0.099 \\
 & 0.8 & 2.088 & 2.089 & 2.094 & 0.021 & 0.022 \\
 \bottomrule
\end{tabular}

\label{tab:my-table}
\end{table*}

\begin{table*}[]
\centering
\caption{Performance comparison between $U_o$ and $U_s$ estimators for convolutional neural network on CIFAR10. The NLL results correspond to the case of distilling the posterior predictive distribution while the MAE on entropy results correspond to the case of distilling the expectation of predictive entropy.}
\begin{tabular}{cccccc}
\toprule
\begin{tabular}[c]{@{}c@{}}Num. training \\ samples\end{tabular} & \begin{tabular}[c]{@{}c@{}}NLL\\ (Teacher)\end{tabular} & \begin{tabular}[c]{@{}c@{}}NLL\\ (Student, $U_o$)\end{tabular} & \begin{tabular}[c]{@{}c@{}}NLL\\ (Student, $U_s$)\end{tabular} & \begin{tabular}[c]{@{}c@{}}MAE \\ (Entropy, $U_o$)\end{tabular} & \begin{tabular}[c]{@{}c@{}}MAE \\ (Entropy, $U_s$)\end{tabular} \\ \midrule
10000 & 0.912 & 1.372 & 1.391 & 0.144 & 0.192 \\ %\hline
\midrule
20000 & 0.798 & 1.184 & 1.179 & 0.210 & 0.231 \\ %\hline
\midrule
50000 & 0.671 & 0.924 & 0.932 & 0.245 & 0.290 \\ %\hline
\bottomrule
\end{tabular}

\label{tab:estimator_comparison_cifar_cnn}
\end{table*}

% \begin{figure}[htbp]
%     \centering
%     \subfigure[]{\includegraphics[width=0.31\textwidth]{figures/cifar_cnn/cifar_cnn_masking_teacher_nll_results.pdf}}
%     \subfigure[]{\includegraphics[width=0.31\textwidth]{figures/cifar_cnn/cifar_cnn_masking_delta_nll_results.pdf}}
%     \subfigure[]{\includegraphics[width=0.31\textwidth]{figures/cifar_cnn/cifar_cnn_masking_delta_entropy_results.pdf}}
%     \caption{Distillation performance while using CNNs on augmented CIFAR10}
%     \label{fig:cnn_mnist_robustness_results}
% \end{figure} 

\subsection{Supplemental Results for Experiment 3: Towards Student  Model Architecture Search} 
\begin{figure*}[htbp]
    \centering
    \subfigure[]{\includegraphics[width=0.24\textwidth]{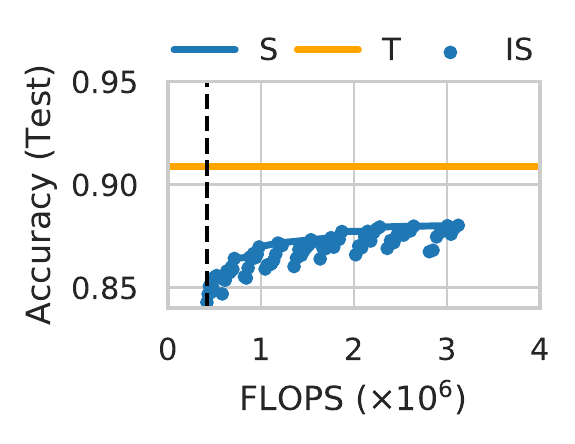}}
    \subfigure[]{\includegraphics[width=0.24\textwidth]{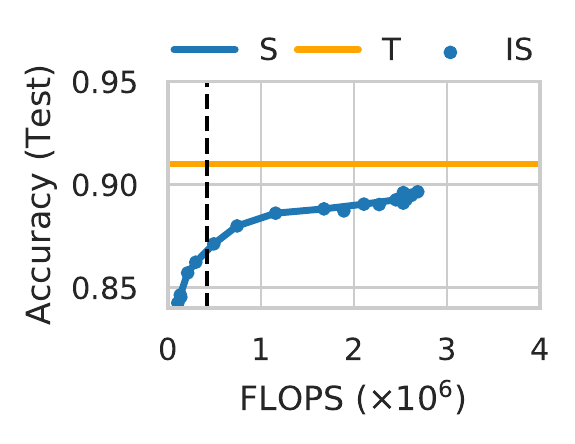}}
    \subfigure[]{\includegraphics[width=0.24\textwidth]{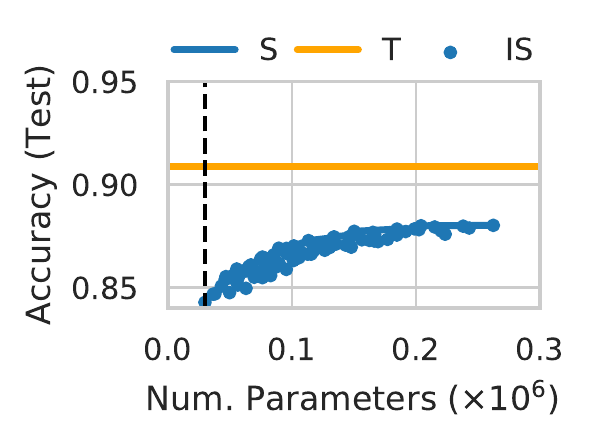}}
    \subfigure[]{\includegraphics[width=0.24\textwidth]{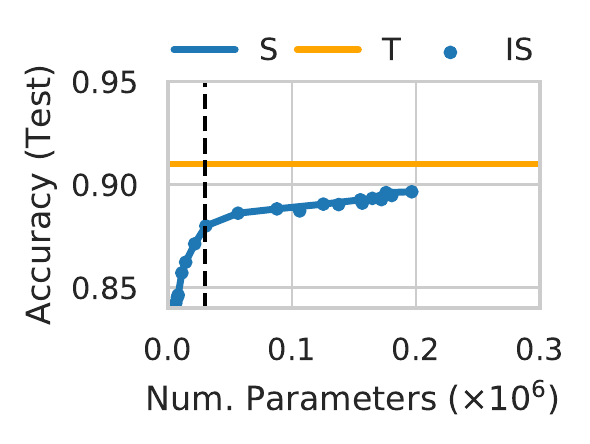}}
    \caption{Accuracy-Storage-Computation tradeoff while using CNNs on MNIST with masking rate $29\%$. (a) Test accuracy using posterior predictive distribution vs FLOPS found using exhaustive search. (b) Test accuracy using posterior predictive distribution vs FLOPS found using group $\ell_1/\ell_2$ with pruning.  (c) Test accuracy using posterior predictive distribution vs storage found using exhaustive search. (d) Test accuracy using posterior predictive distribution vs storage found using group $\ell_1/\ell_2$ with pruning. The optimal student model for this configuration is obtained with group $\ell_1/\ell_2$ pruning. It has approximately $6.6 \times$ the number of parameters and $6.4 \times$ the FLOPS of the base student model. Notation: ``S" - pareto frontier of the student models, ``T" - Teacher, ``IS" - Individual Student. The black dashed line denotes the FLOPS/no. of parameters of the base student model having the same architecture as a teacher model.}
    \label{fig:cnn_accuracy_tradeoff_results}
\end{figure*}

\begin{figure*}[htbp]
    \centering
    \subfigure[]{\includegraphics[width=0.24\textwidth]{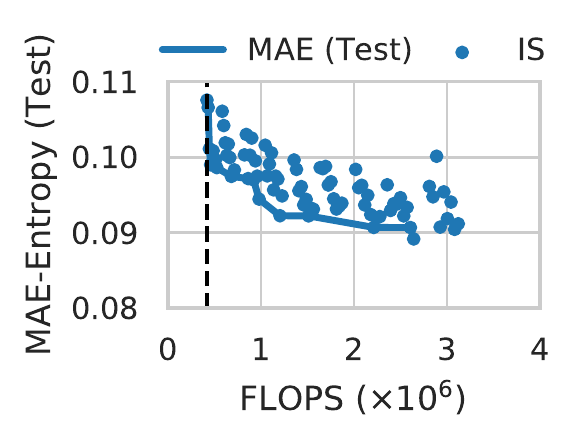}}
    \subfigure[]{\includegraphics[width=0.24\textwidth]{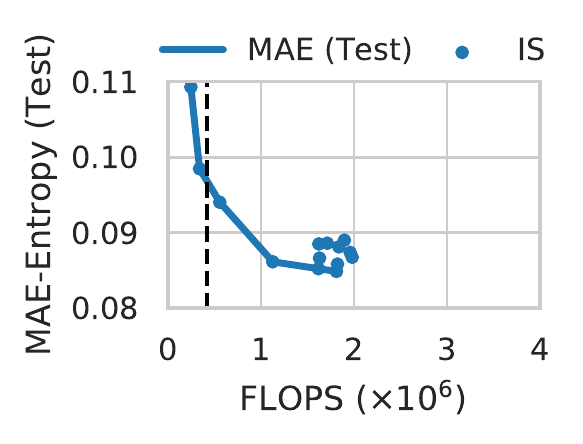}}
    \subfigure[]{\includegraphics[width=0.24\textwidth]{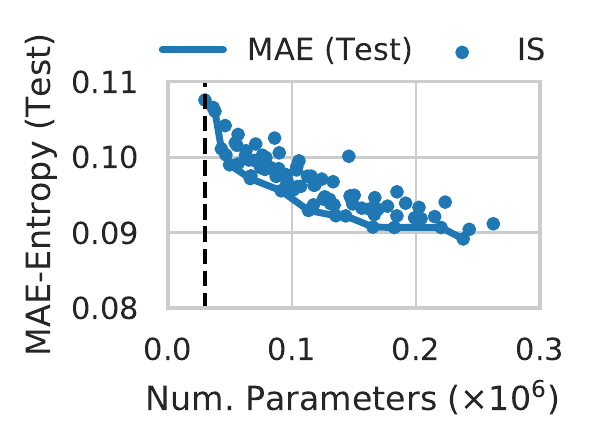}}
    \subfigure[]{\includegraphics[width=0.24\textwidth]{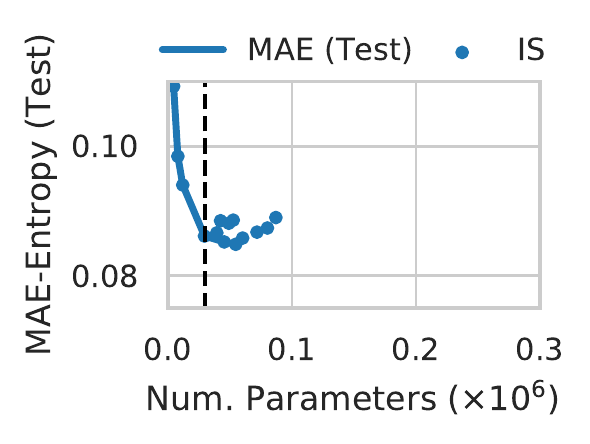}}
    \caption{Entropy Error-Storage-Computation tradeoff while using CNNs on MNIST with masking rate $29\%$. (a) Test mean absolute error for posterior entropy vs FLOPS found using exhaustive search. (b)  Test mean absolute error for posterior entropy vs FLOPS found using group $\ell_1/\ell_2$ with pruning.  (c)  Test mean absolute error for posterior entropy vs storage found using exhaustive search. (d)  Test mean absolute error for posterior entropy vs storage found using group $\ell_1/\ell_2$ with pruning. The optimal student model for this configuration is obtained with group $\ell_1/\ell_2$ pruning. It has approximately $1.8 \times$ the number of parameters and $4.3 \times$ the FLOPS of the base student model. Notation: ``MAE (Test)" - pareto frontier of the MAEs obtained using different student models, ``IS" - Individual Student. The black dashed line denotes the FLOPS/no. of parameters of the base student model having the same architecture as a teacher model.}
    \label{fig:cnn_entropy_tradeoff_results}
\end{figure*}

\begin{figure*}[htbp]
    \centering
    \subfigure[]{\includegraphics[width=0.24\textwidth]{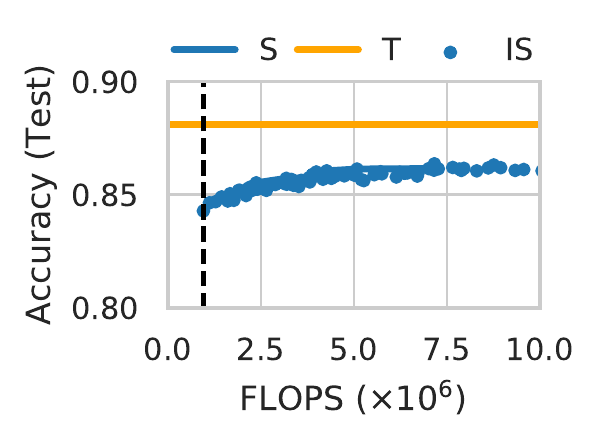}}
    \subfigure[]{\includegraphics[width=0.24\textwidth]{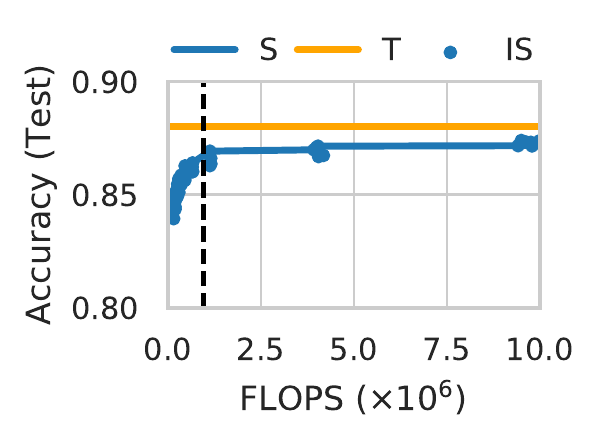}}
    \subfigure[]{\includegraphics[width=0.24\textwidth]{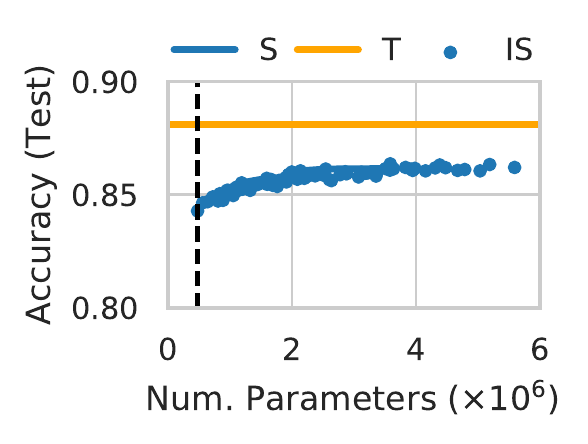}}
    \subfigure[]{\includegraphics[width=0.24\textwidth]{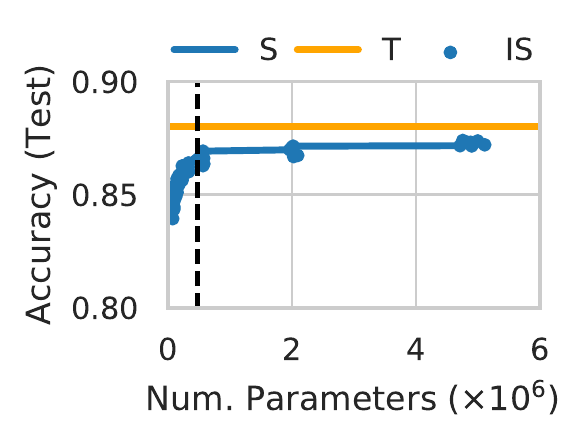}}
    \caption{Accuracy-Storage-Computation tradeoff while using Fully-connected networks on MNIST with masking rate $29\%$. (a) Test accuracy using posterior predictive distribution vs FLOPS found using exhaustive search. (b) Test accuracy using posterior predictive distribution vs FLOPS found using group $\ell_1/\ell_2$ with pruning.  (c) Test accuracy using posterior predictive distribution vs storage found using exhaustive search. (d) Test accuracy using posterior predictive distribution vs storage found using group $\ell_1/\ell_2$ with pruning. The optimal student model for this configuration is obtained with group $\ell_1/\ell_2$ pruning. It has approximately $9.9 \times$ the number of parameters and $10 \times$ the FLOPS of the base student model. Notation: ``S" - pareto frontier of the student models, ``T" - Teacher, ``IS" - Individual Student. The black dashed line denotes the FLOPS/no. of parameters of the base student model having the same architecture as a teacher model.}
    \label{fig:fcnn_accuracy_tradeoff_results}
\end{figure*}

\begin{figure*}[htbp]
    \centering
    \subfigure[]{\includegraphics[width=0.24\textwidth]{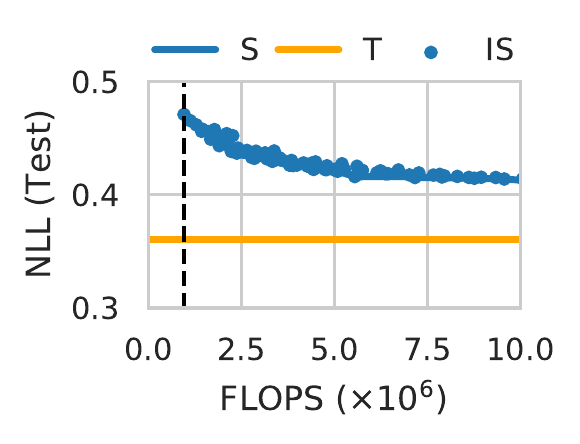}}
    \subfigure[]{\includegraphics[width=0.24\textwidth]{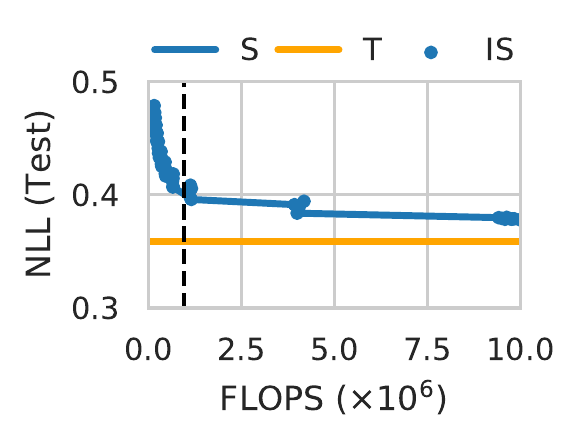}}
    \subfigure[]{\includegraphics[width=0.24\textwidth]{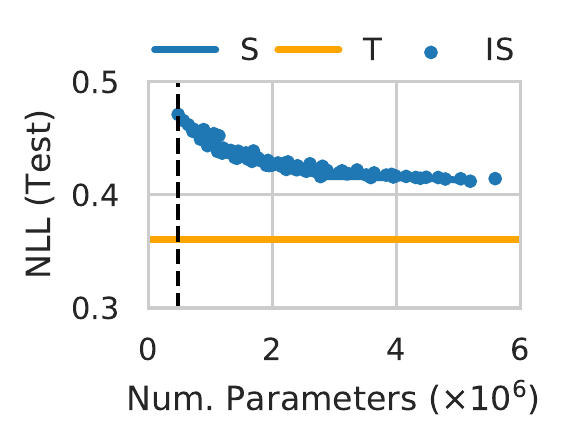}}
    \subfigure[]{\includegraphics[width=0.24\textwidth]{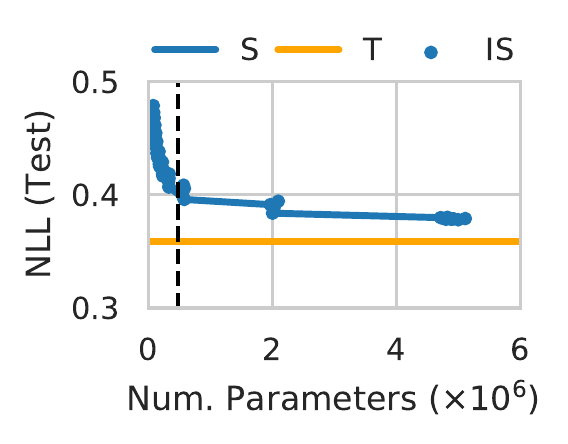}}
    \caption{NLL-Storage-Computation tradeoff while using Fully-connected networks on MNIST with masking rate $29\%$. (a) Test negative log likelihood of posterior predictive distribution vs FLOPS found using exhaustive search. (b) Test negative log likelihood of posterior predictive distribution vs FLOPS found using group $\ell_1/\ell_2$ with pruning.  (c) Test negative log likelihood of posterior predictive distribution vs storage found using exhaustive search. (d) Test negative log likelihood of posterior predictive distribution vs storage found using group $\ell_1/\ell_2$ with pruning. The optimal student model for this configuration is obtained with group $\ell_1/\ell_2$ pruning. It has approximately $9.9 \times$ the number of parameters and $10 \times$ the FLOPS of the base student model. Notation: ``S" - pareto frontier of the student models, ``T" - Teacher, ``IS" - Individual Student. The black dashed line denotes the FLOPS/no. of parameters of the base student model having the same architecture as a teacher model.}
    \label{fig:fcnn_nll_tradeoff_results}
\end{figure*}

\begin{figure*}[htbp]
    \centering
    \subfigure[]{\includegraphics[width=0.24\textwidth]{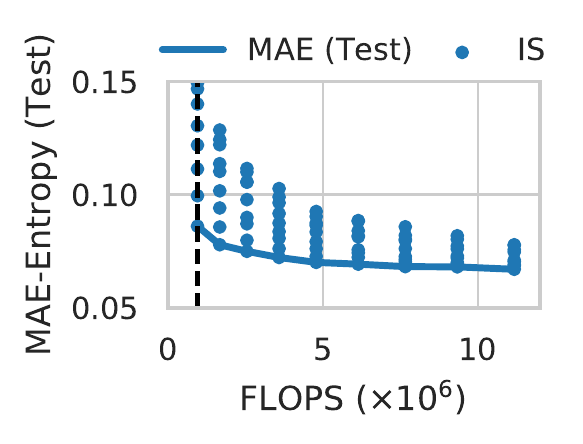}}
    \subfigure[]{\includegraphics[width=0.24\textwidth]{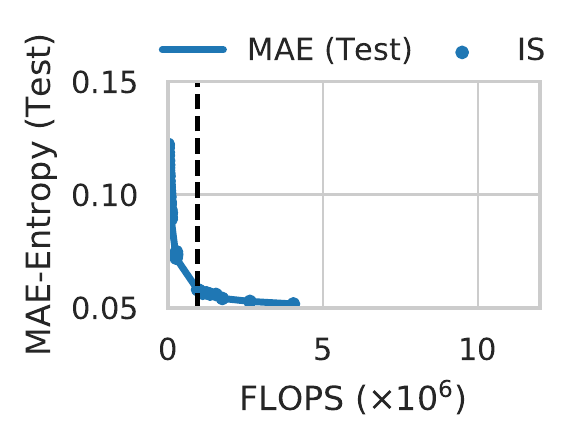}}
    \subfigure[]{\includegraphics[width=0.24\textwidth]{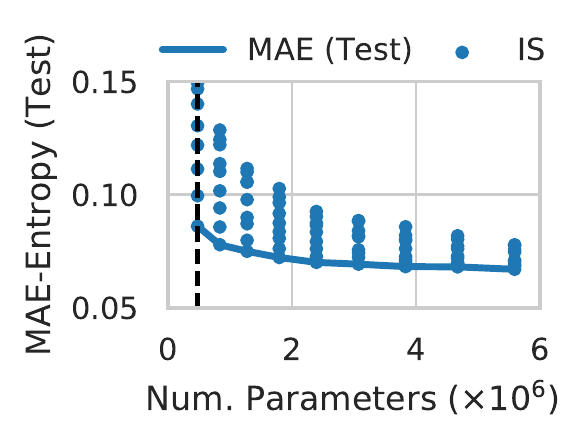}}
    \subfigure[]{\includegraphics[width=0.24\textwidth]{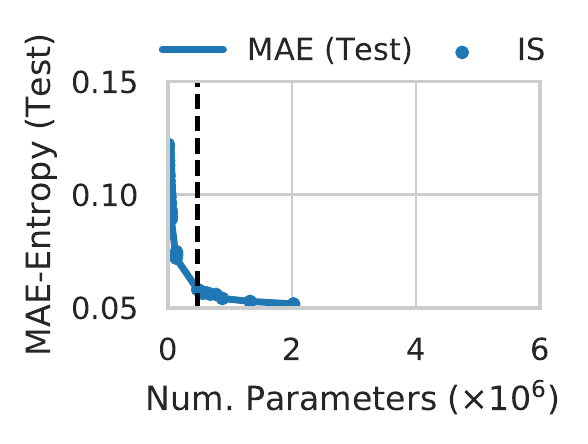}}
    \caption{Entropy Error-Storage-Computation tradeoff while using Fully-connected networks on MNIST with masking rate $29\%$. (a) Test mean absolute error for posterior entropy vs FLOPS found using exhaustive search. (b)  Test mean absolute error for posterior entropy vs FLOPS found using group $\ell_1/\ell_2$ with pruning.  (c)  Test mean absolute error for posterior entropy vs storage found using exhaustive search. (d)  Test mean absolute error for posterior entropy vs storage found using group $\ell_1/\ell_2$ with pruning. The optimal student model for this configuration is obtained with group $\ell_1/\ell_2$ pruning. It has approximately $4.2 \times$ the number of parameters and $4.2 \times$ the FLOPS of the base student model. Notation: ``MAE (Test)" - pareto frontier of the MAEs obtained using different student models, ``IS" - Individual Student. The black dashed line denotes the FLOPS/no. of parameters of the base student model having the same architecture as a teacher model.}
    \label{fig:fcnn_entropy_tradeoff_results}
\end{figure*}

\begin{figure*}[htbp]
    \centering
    \subfigure[]{\includegraphics[width=0.24\textwidth]{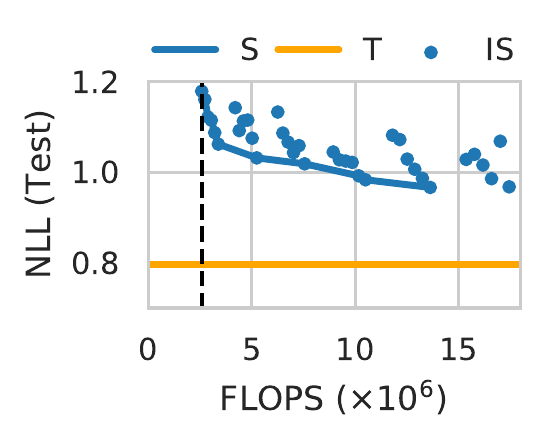}}
    \subfigure[]{\includegraphics[width=0.24\textwidth]{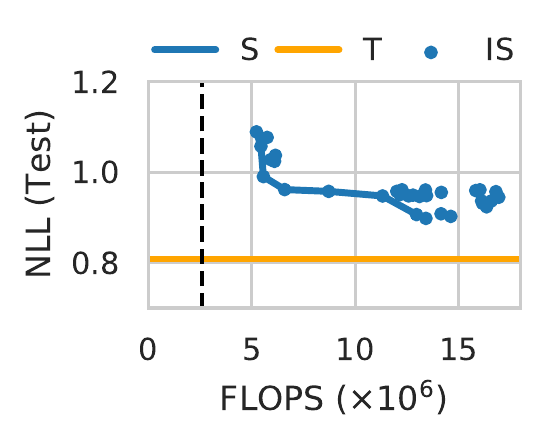}}
    \subfigure[]{\includegraphics[width=0.24\textwidth]{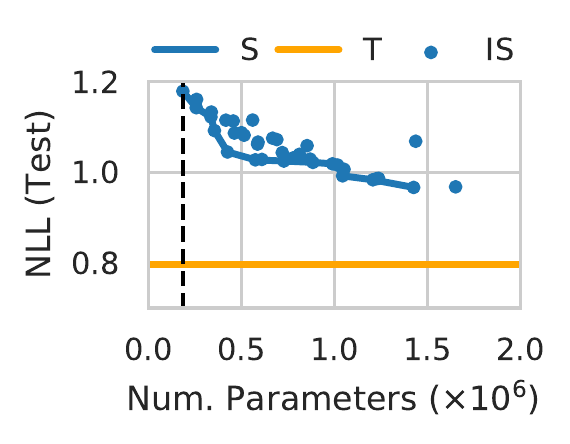}}
    \subfigure[]{\includegraphics[width=0.24\textwidth]{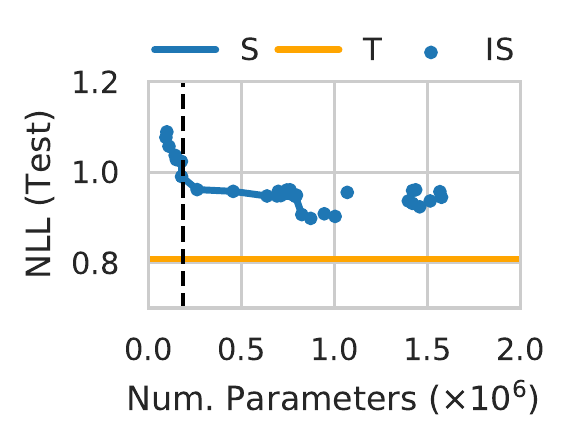}}
     \caption{NLL-Storage-Computation tradeoff while using CNNs on CIFAR10 with training set size of 20,000 samples. (a) Test negative log likelihood of posterior predictive distribution vs FLOPS found using exhaustive search. (b) Test negative log likelihood of posterior predictive distribution vs FLOPS found using group $\ell_1/\ell_2$ with pruning.  (c) Test negative log likelihood of posterior predictive distribution vs storage found using exhaustive search. (d) Test negative log likelihood of posterior predictive distribution vs storage found using group $\ell_1/\ell_2$ with pruning. The optimal student model for this configuration is obtained with group $\ell_1/\ell_2$ pruning. It has approximately $ 4.7 \times$ the number of parameters and $5.2 \times$ the FLOPS of the base student model. Notation: ``S" - pareto frontier of the student models, ``T" - Teacher, ``IS" - Individual Student. The black dashed line denotes the FLOPS/no. of parameters of the base student model having the same architecture as a teacher model.}
    \label{fig:cifar_cnn_nll_tradeoff_results}
\end{figure*}

\begin{figure*}[htbp]
    \centering
    \subfigure[]{\includegraphics[width=0.24\textwidth]{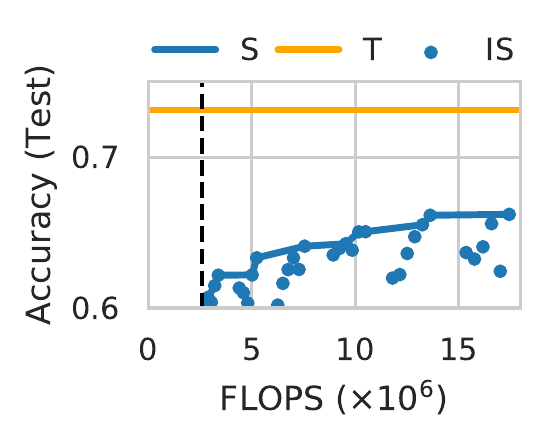}}
    \subfigure[]{\includegraphics[width=0.24\textwidth]{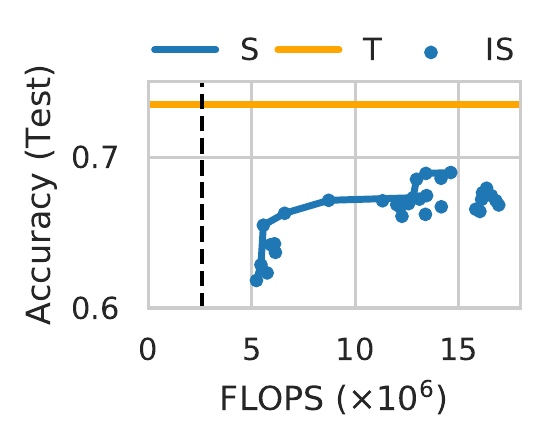}}
    \subfigure[]{\includegraphics[width=0.24\textwidth]{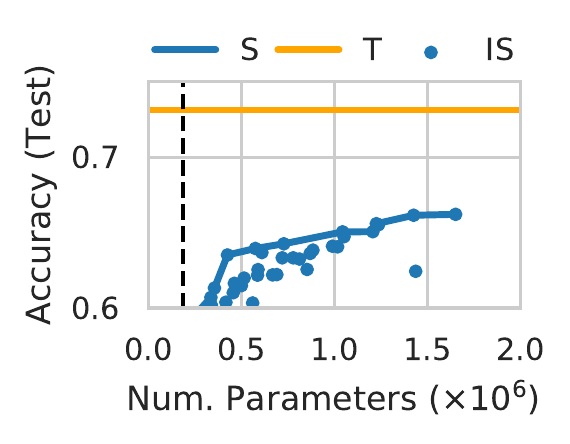}}
    \subfigure[]{\includegraphics[width=0.24\textwidth]{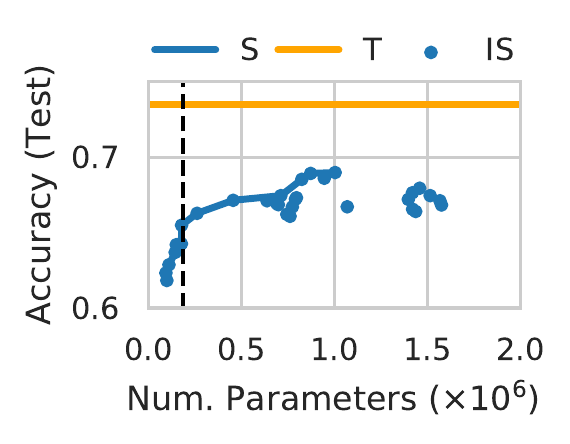}}
    \caption{Accuracy-Storage-Computation tradeoff while using CNNs on CIFAR10 with sub-sampling training data to 20,000 samples. (a) Test accuracy using posterior predictive distribution vs FLOPS found using exhaustive search. (b) Test accuracy using posterior predictive distribution vs FLOPS found using group $\ell_1/\ell_2$ with pruning.  (c) Test accuracy using posterior predictive distribution vs storage found using exhaustive search. (d) Test accuracy using posterior predictive distribution vs storage found using group $\ell_1/\ell_2$ with pruning. The optimal student model for this configuration is obtained with group $\ell_1/\ell_2$ pruning. It has approximately $ 5.4 \times$ the number of parameters and $5.6 \times$ the FLOPS of the base student model. Notation: ``S" - pareto frontier of the student models, ``T" - Teacher, ``IS" - Individual Student. The black dashed line denotes the FLOPS/no. of parameters of the base student model having the same architecture as a teacher model.}
    \label{fig:cifar_cnn_accuracy_tradeoff_results}
\end{figure*}

\begin{figure*}[htbp]
    \centering
    \subfigure[]{\includegraphics[width=0.24\textwidth]{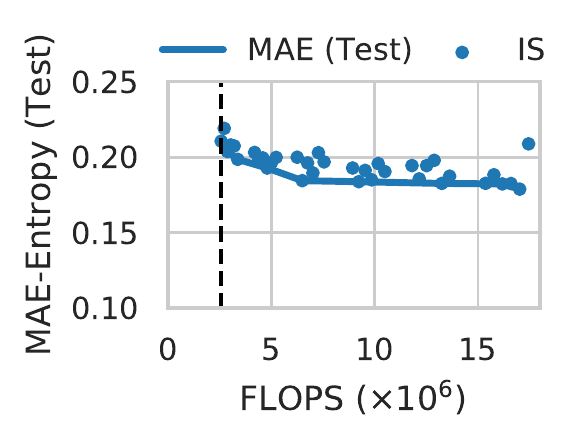}}
    \subfigure[]{\includegraphics[width=0.24\textwidth]{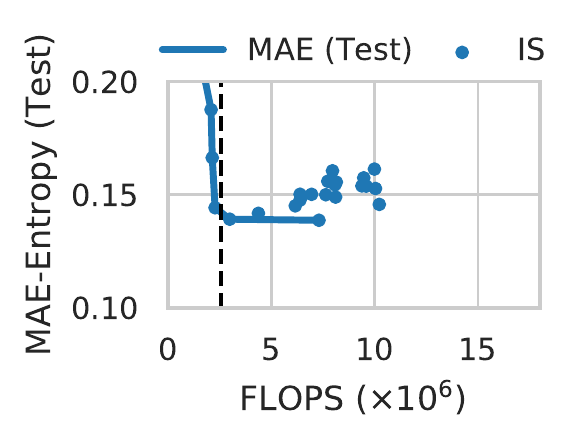}}
    \subfigure[]{\includegraphics[width=0.24\textwidth]{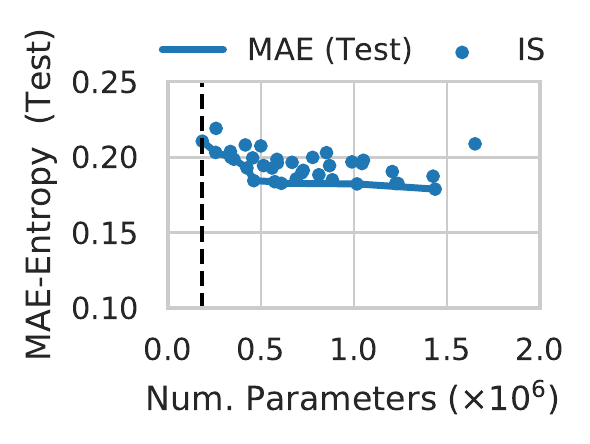}}
    \subfigure[]{\includegraphics[width=0.24\textwidth]{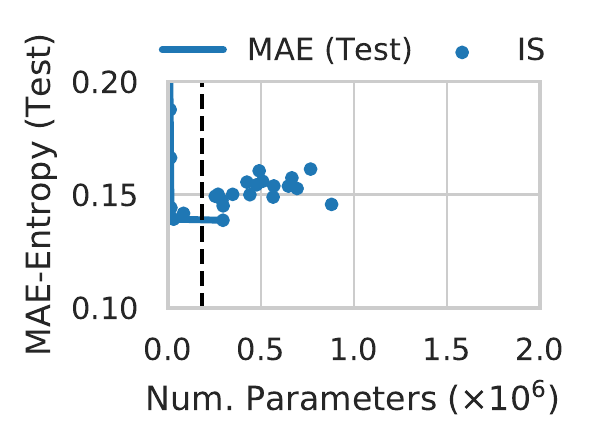}}
    \caption{Entropy Error-Storage-Computation tradeoff while using CNNs on CIFAR10 with sub-sampling training data to 20,000 samples. (a) Test mean absolute error for posterior entropy vs FLOPS found using exhaustive search. (b)  Test mean absolute error for posterior entropy vs FLOPS found using group $\ell_1/\ell_2$ with pruning.  (c)  Test mean absolute error for posterior entropy vs storage found using exhaustive search. (d)  Test mean absolute error for posterior entropy vs storage found using group $\ell_1/\ell_2$ with pruning. The optimal student model for this configuration is obtained with group $\ell_1/\ell_2$ pruning. It has approximately $1.6 \times$ the number of parameters and $ 2.8 \times$ the FLOPS of the base student model. Notation: ``MAE (Test)" - pareto frontier of the MAEs obtained using different student models, ``IS" - Individual Student. The black dashed line denotes the FLOPS/no. of parameters of the base student model having the same architecture as a teacher model.}
    \label{fig:cifar_cnn_entropy_tradeoff_results}
\end{figure*}

% In the previous experiments, we saw that using a student model as the same architecture of teacher leads to performance gaps in moderate levels of uncertainty. In this set of experiments, we show that increasing the capacity of the student model helps in mitigating some of the performance gaps we saw in the earlier section. For each configuration of model and dataset, we examine different architectures by tweaking the values of $K_1$ and $K_2$ for the student models. This is equivalent to doing a grid-search over the different possible configurations of a given model.  As a comparison, we look at starting with overparameterised student models whose architecture is given by highest value of $(K_1, K_2)$ from our grid, and pruning them back using Group Lasso regularization. We run distillation against different values of the regularization strength $\lambda_{R}$ to ultimately obtain models of different sizes. For comparing the group lasso regularization approach against grid-search, we look at plotting performance against number of parameters (storage cost) and computation cost (FLOPS). FLOPS indicate the number of floating point operations required for a single forward pass through the model for a given input.
The additional results from running Experiment 3 (Section 4.4) on different combinations of model type, dataset, and performance metrics have been given in Figures[\ref{fig:cnn_accuracy_tradeoff_results} - \ref{fig:cifar_cnn_entropy_tradeoff_results}].

% \begin{figure}[htbp]
%     \centering
%     \subfigure[]{\includegraphics[width=0.24\textwidth]{figures/mnist_cnn/cnn_nll_flops_exhaustive_results.pdf}}
%     \subfigure[]{\includegraphics[width=0.24\textwidth]{figures/mnist_cnn/cnn_nll_flops_pruning_results.pdf}}
%     \subfigure[]{\includegraphics[width=0.24\textwidth]{figures/mnist_cnn/cnn_nll_num_params_exhaustive_results.pdf}}
%     \subfigure[]{\includegraphics[width=0.24\textwidth]{figures/mnist_cnn/cnn_nll_num_params_pruning_results.pdf}}
%     \caption{NLL-Storage-Computation tradeoff while using CNNs on augmented MNIST}
%     \label{fig:cnn_nll_tradeoff_results}
% \end{figure}    

% \begin{figure}[htbp]
%     \centering
%     \subfigure[]{\includegraphics[width=0.24\textwidth]{figures/cifar_cnn/cifar_cnn_nll_flops_exhaustive_results.pdf}}
%     \subfigure[]{\includegraphics[width=0.24\textwidth]{figures/cifar_cnn/cifar_cnn_nll_flops_pruning_results.pdf}}
%     \subfigure[]{\includegraphics[width=0.24\textwidth]{figures/cifar_cnn/cifar_cnn_nll_num_params_exhaustive_results.pdf}}
%     \subfigure[]{\includegraphics[width=0.24\textwidth]{figures/cifar_cnn/cifar_cnn_nll_num_params_pruning_results.pdf}}
%     \caption{NLL-Storage-Computation tradeoff while using CNNs on CIFAR10}
%     \label{fig:cifar_cnn_nll_tradeoff_results}
% \end{figure}    

% \subsection{Multi-Output Student Models}
% In all the previous experiments, the student model was trained to distill a single posterior expectation quantity. Thus, in cases where we want to obtain estimates of multiple posterior expectations, we need to store as many models and compute as many forward passes through the student as the number of posterior expectations desired. Similar to the previous set of experiments, we produce the performance tradeoff against storage and computation cost for different configurations of models, datasets and tasks.

\subsection{Supplemental Results for Experiment 4: Uncertainty Quantification for Downstream tasks}
\begin{figure*}[htbp]
    \centering
    \subfigure[MNIST-FCNN]{\includegraphics[width=0.45\textwidth]{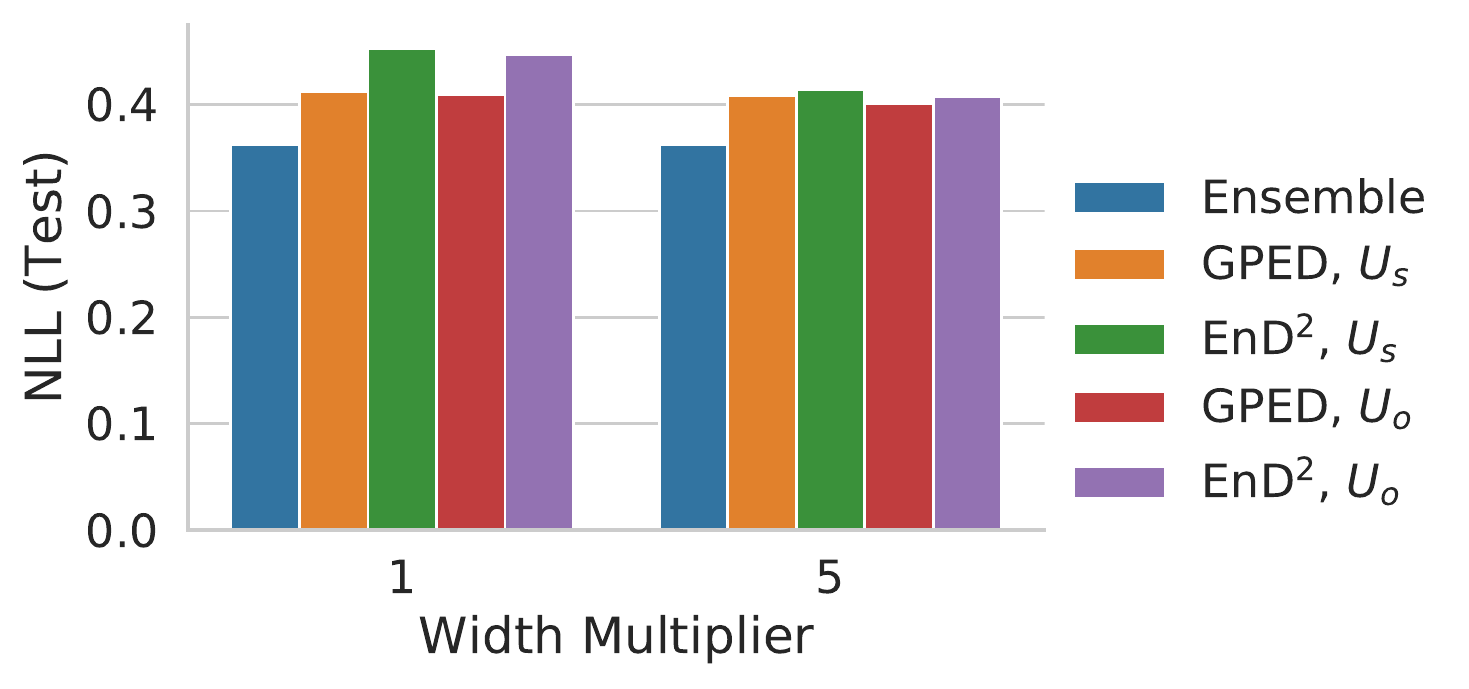}}
    \subfigure[MNIST-FCNN]{\includegraphics[width=0.45\textwidth]{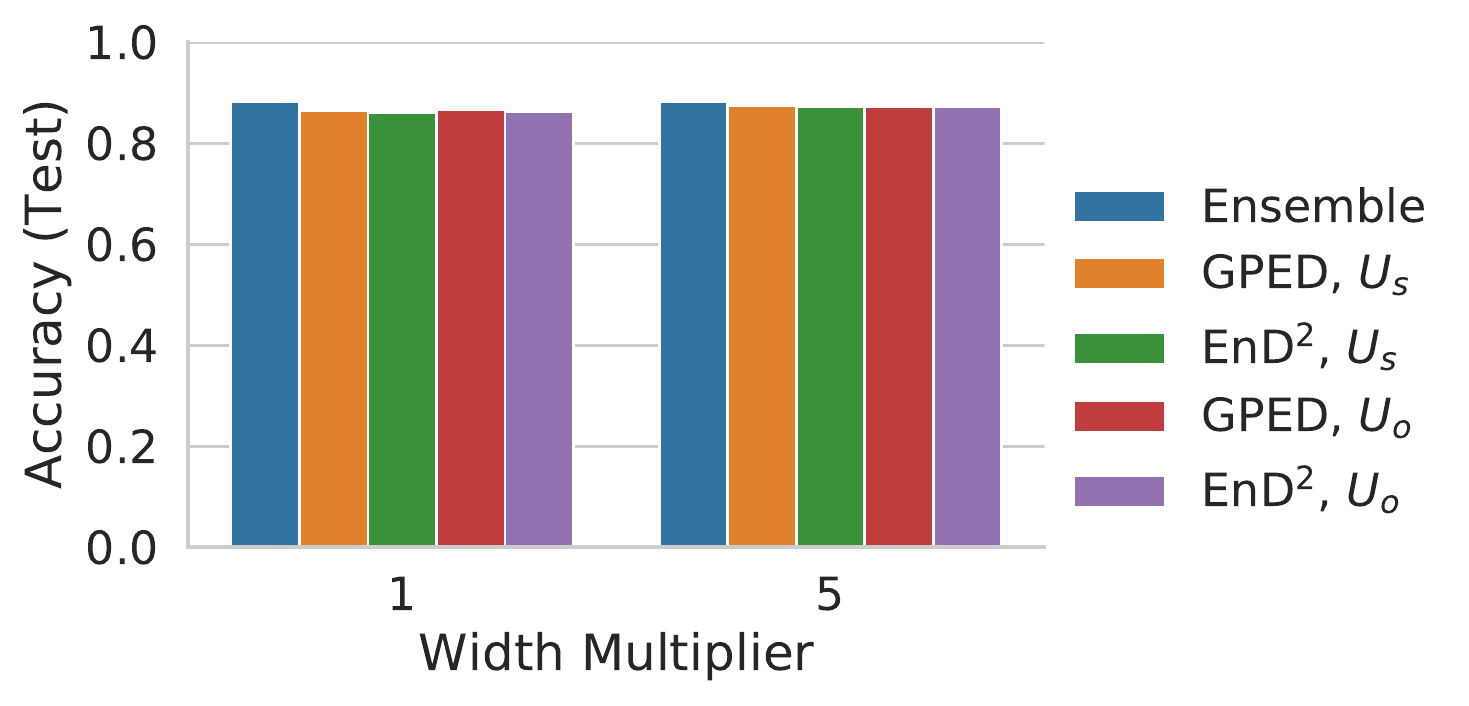}}
    \subfigure[MNIST-CNN]{\includegraphics[width=0.45\textwidth]{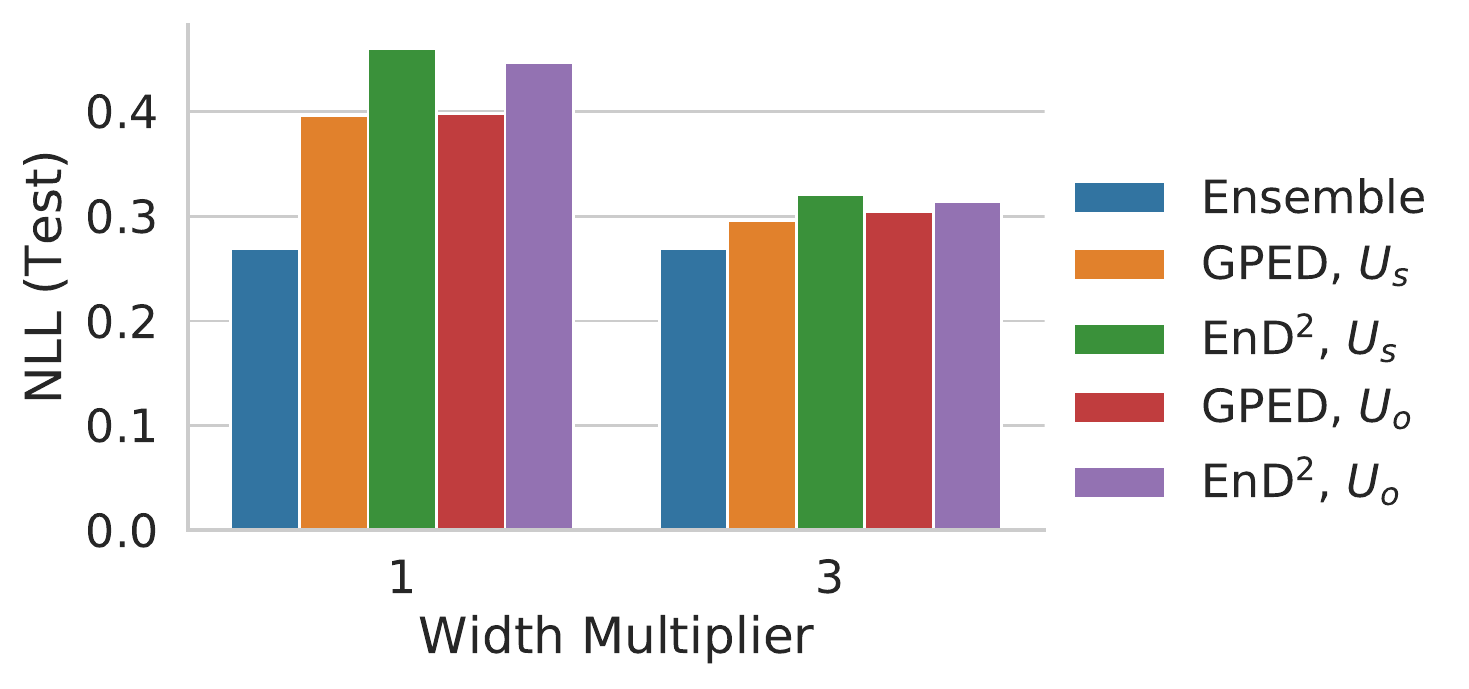}}
    \subfigure[MNIST-CNN]{\includegraphics[width=0.45\textwidth]{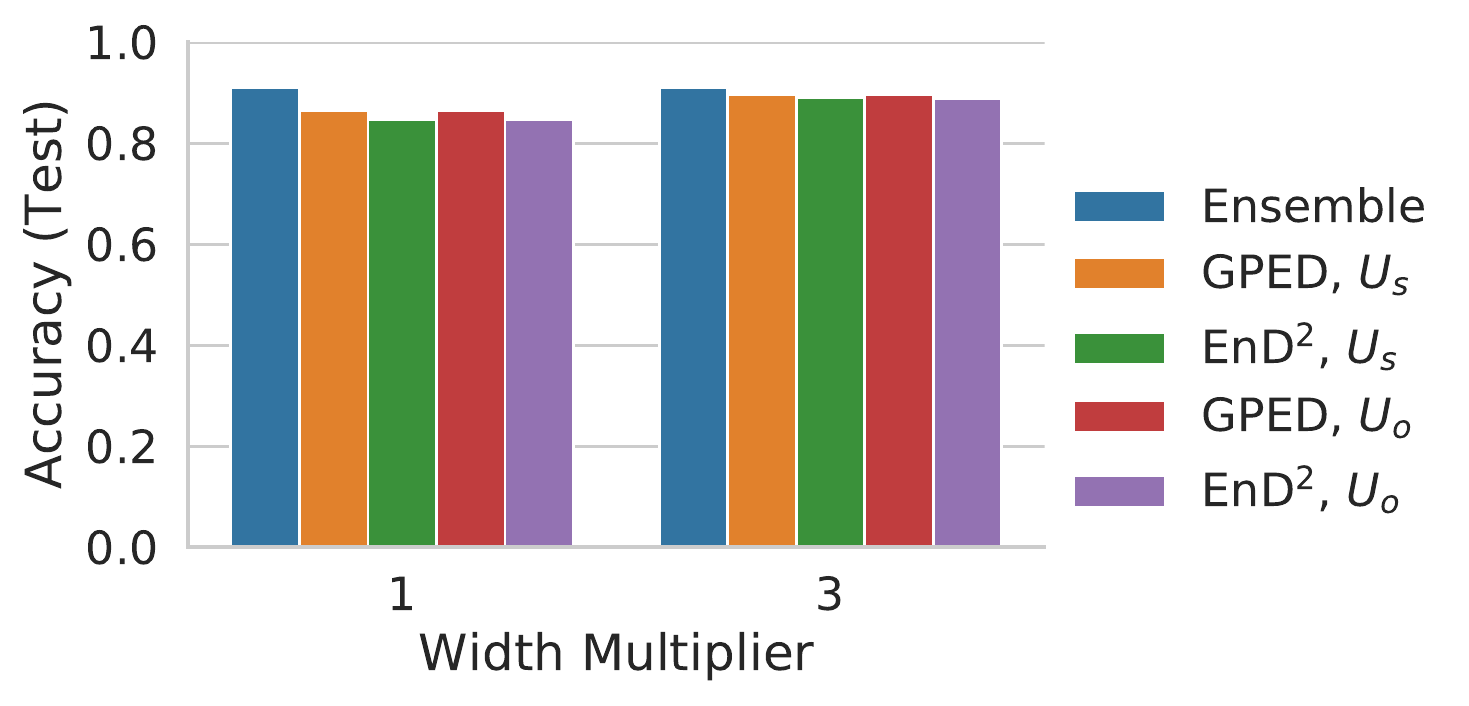}}
    \subfigure[CIFAR10-CNN]{\includegraphics[width=0.45\textwidth]{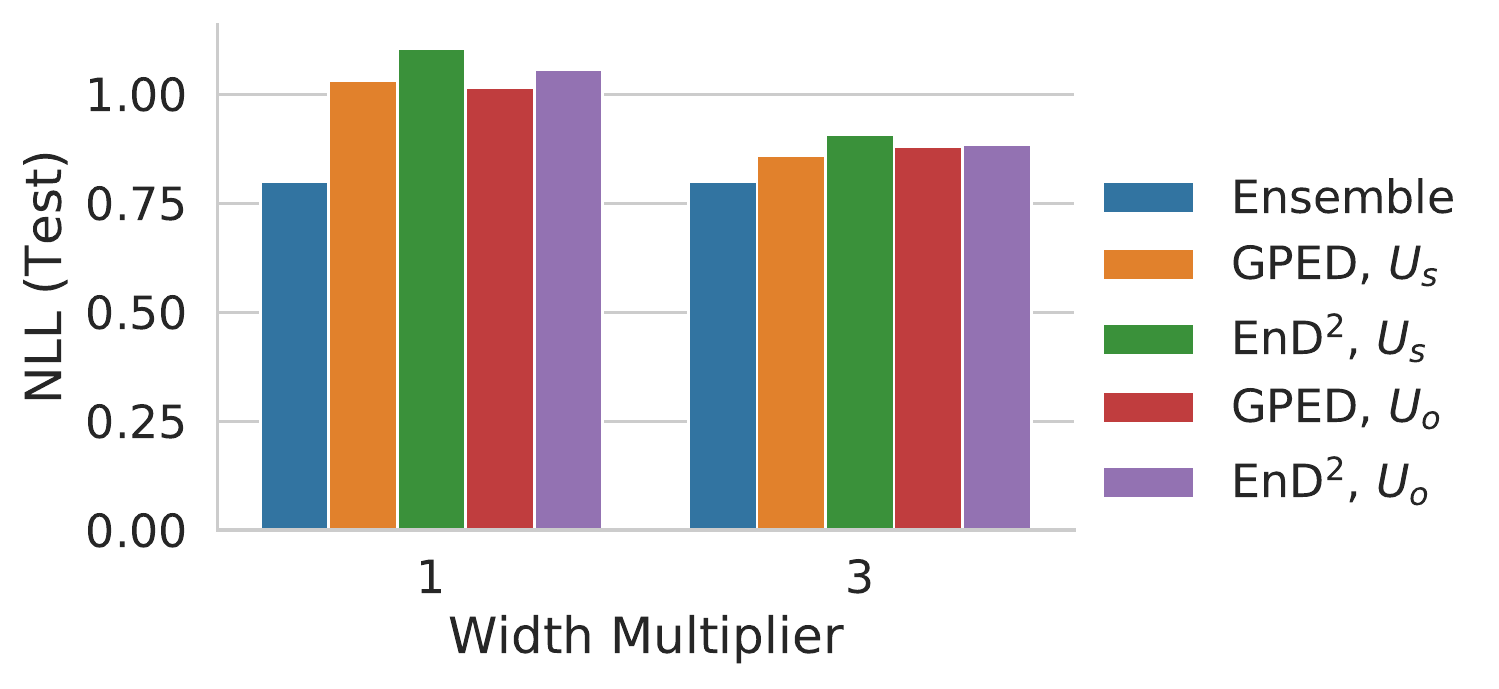}}
    \subfigure[CIFAR10-CNN]{\includegraphics[width=0.45\textwidth]{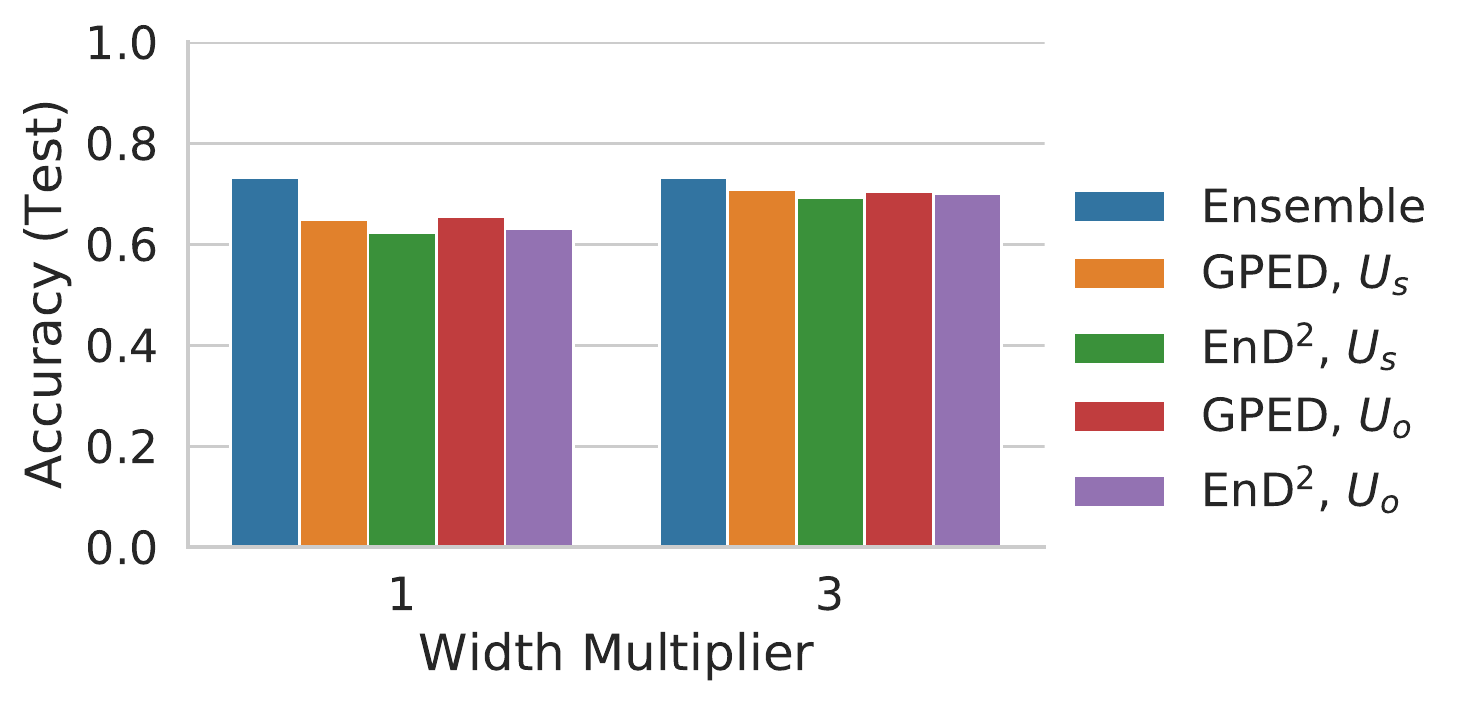}}
    % \subfigure[]{\includegraphics[width=0.33\textwidth]{figures/cifar_cnn/cifar_cnn_accuracy_num_params_exhaustive_results.pdf}}
    \caption{Test accuracy and negative log likelihood comparison between GPED and EnD\textsuperscript{2} for different dataset-model-estimator combinations.}
    \label{fig:gped_pn_comparison}
\end{figure*}

\begin{figure*}[htbp]
    \centering
    \subfigure[MNIST-FCNN]{\includegraphics[width=0.3\linewidth]{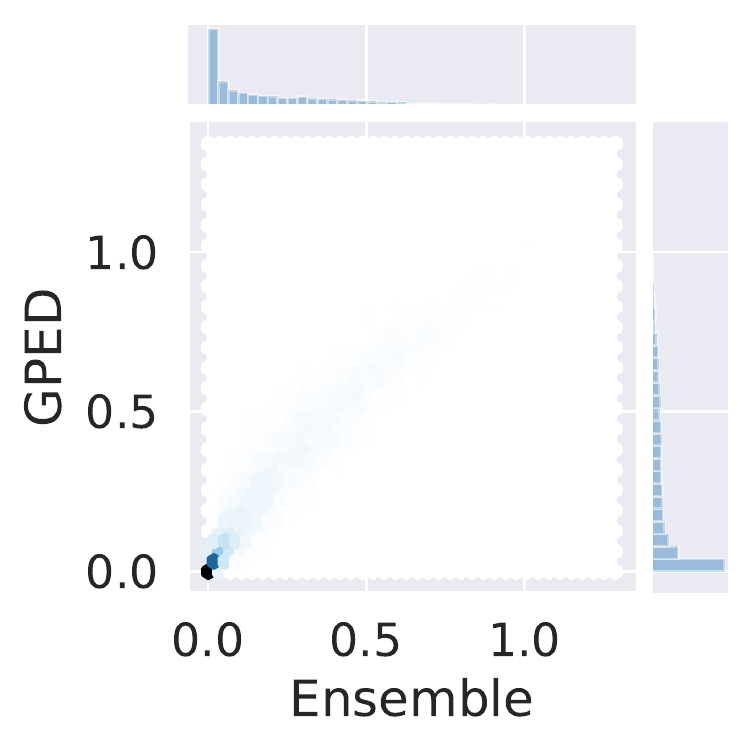}}
    \subfigure[MNIST-FCNN]{\includegraphics[width=0.3\linewidth]{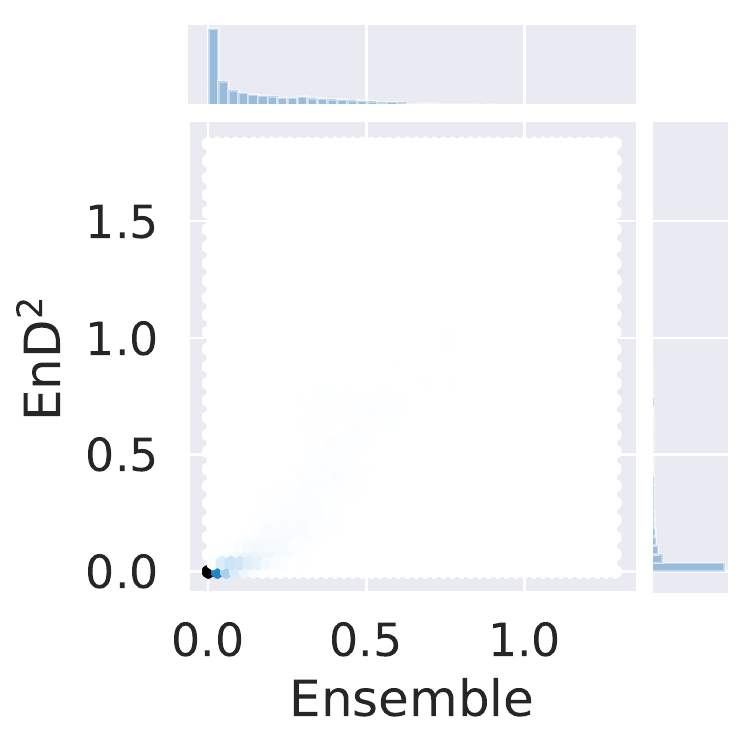}}
    \subfigure[MNIST-CNN]{\includegraphics[width=0.3\linewidth]{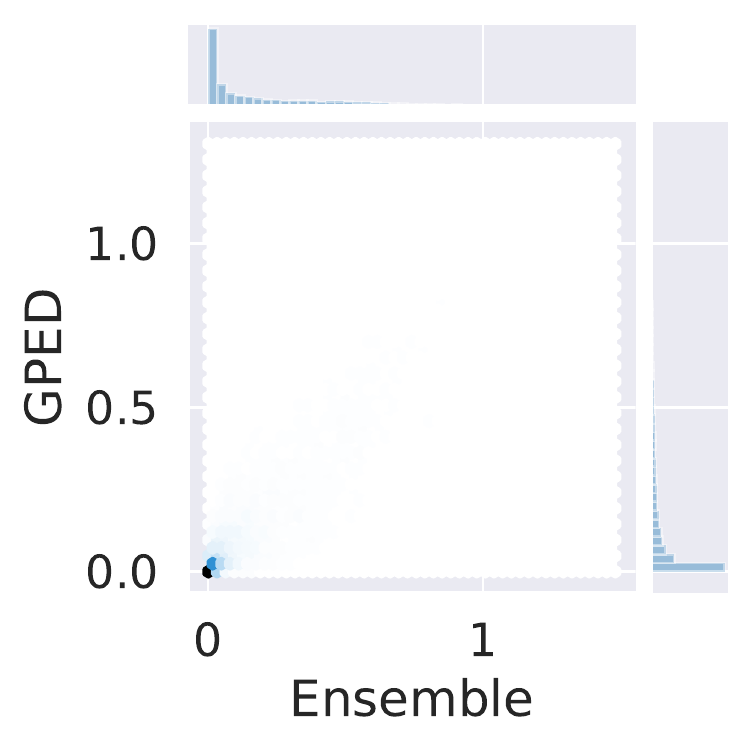}}\\
    \subfigure[MNIST-CNN]{\includegraphics[width=0.3\linewidth]{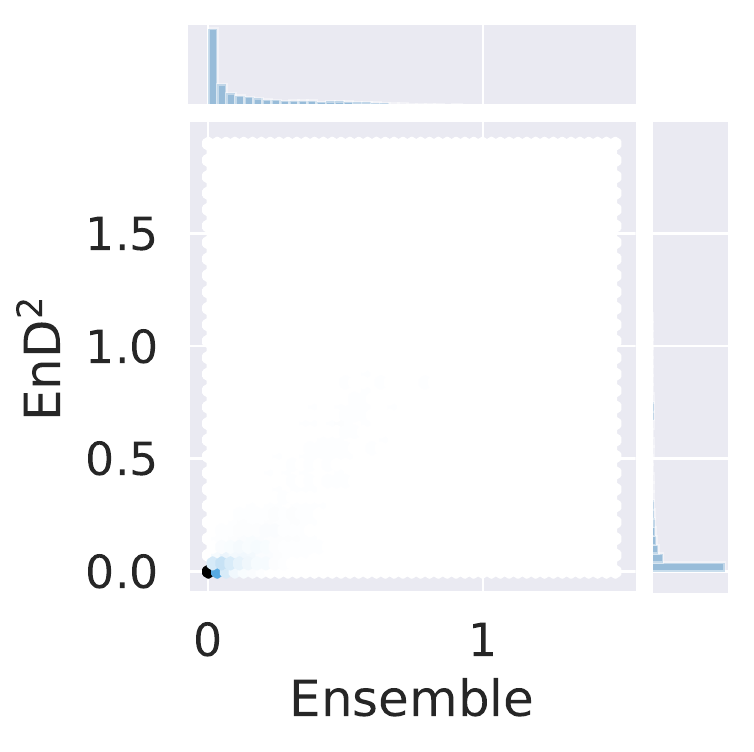}}
    \subfigure[CIFAR10-CNN]{\includegraphics[width=0.3\linewidth]{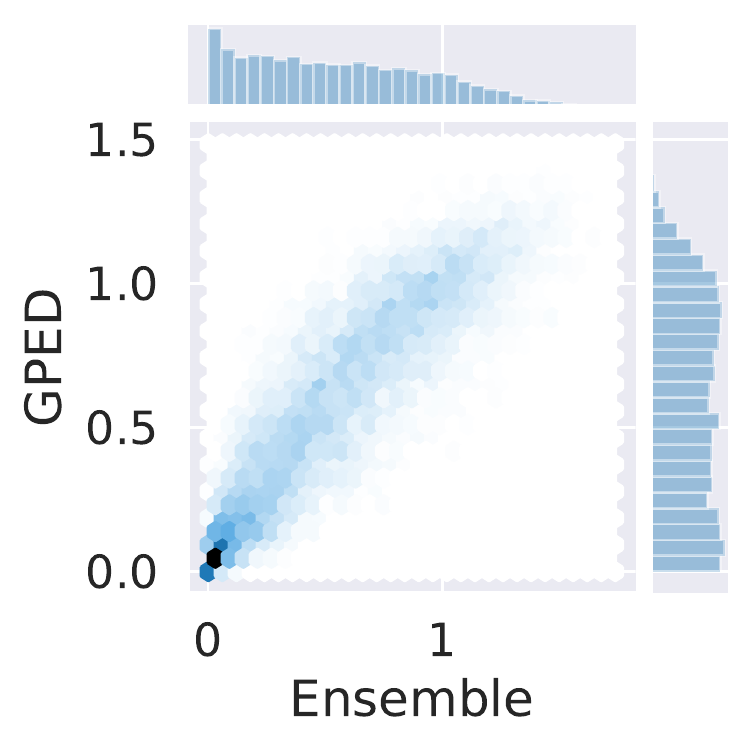}}
    \subfigure[CIFAR10-CNN]{\includegraphics[width=0.3\linewidth]{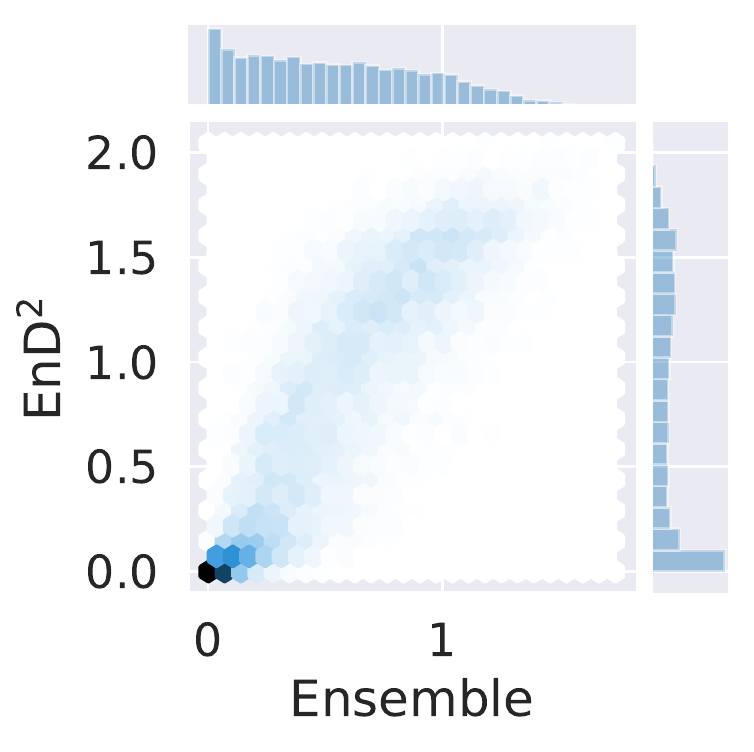}}
    % \subfigure[]{\includegraphics[width=0.33\textwidth]{figures/cifar_cnn/cifar_cnn_accuracy_num_params_exhaustive_results.pdf}}
    \caption{Joint and marginal distributions for expected data uncertainty (also known as the expected entropy) by using $U_s$ estimator for EnD\textsuperscript{2} and using the largest model obtained using width multiplier. The expected data entropy is over in-distribution test dataset. An ideal distillation approach would match the marginal distribution of the teacher ensemble given on the top of each plot, as well as have the joint density concentrated on the diagonal. Based on this properties, it is evident that GPED does a better job at tracking the expected data uncertainty.}
    \label{fig:joint_distribution_plots}
\end{figure*}

Additional results on performance comparison between GPED and EnD\textsuperscript{2} on in-distribution test datasets have been given in Tables [\ref{tab:test-metrics-comparison-us-small-model} - \ref{tab:test-metrics-comparison-uo-large-model}] and Figure \ref{fig:gped_pn_comparison}. As an illustration, we also present the joint \& distribution plots in Figure \ref{fig:joint_distribution_plots} for distilled expected entropy (expected data uncertainty) pertaining to Table \ref{tab:test-metrics-comparison-us-large-model} to show how our GPED and EnD\textsuperscript{2} compare against the Bayesian ensemble.

The supplemental results for OOD detection are given in Tables [\ref{tab:ood-auroc-us-small-model}-\ref{tab:ood-auroc-uo-large-model}] and the supplemental results on nDCG scores are given in Tables [\ref{tab:ndcg-comparison-us-small-model} - \ref{tab:ndcg-comparison-u0-large-model}].

\begin{table}[htbp]
\centering
\caption{In-distribution Test set metrics comparison using $U_s$ and base student model matching the teacher architecture.}
\label{tab:test-metrics-comparison-us-small-model}
\resizebox{\linewidth}{!}{
\begin{tabular}{cccccc}
\toprule
\begin{tabular}[c]{@{}c@{}}Model/\\ Dataset\end{tabular} &
  \begin{tabular}[c]{@{}c@{}}NLL\\ (Ensemble)\end{tabular} &
  \begin{tabular}[c]{@{}c@{}}NLL\\ (GPED)\end{tabular} &
  \begin{tabular}[c]{@{}c@{}}NLL\\ (EnD\textsuperscript{2})\end{tabular} &
  \begin{tabular}[c]{@{}c@{}}MAE \\ Entropy\\ (GPED)\end{tabular} &
  \begin{tabular}[c]{@{}c@{}}MAE \\Entropy\\ (EnD\textsuperscript{2})\end{tabular} \\ \midrule
\begin{tabular}[c]{@{}c@{}}FCNN/\\ MNIST\end{tabular}  & 0.362 & 0.412 & 0.452 & 0.063 & 0.113 \\ \midrule
\begin{tabular}[c]{@{}c@{}}CNN/\\ MNIST\end{tabular}   & 0.269 & 0.396 & 0.460 & 0.121 & 0.175 \\ \midrule
\begin{tabular}[c]{@{}c@{}}CNN/\\ CIFAR10\end{tabular} & 0.799 & 1.032 & 1.104 & 0.181 & 0.424 \\ \bottomrule
\end{tabular}
}
\end{table} 

\begin{table}[htbp]
\centering
\caption{In-distribution Test set metrics comparison using $U_o$ and base student model matching the teacher architecture.}
\label{tab:test-metrics-comparison-uo-small-model}
\resizebox{\linewidth}{!}{
\begin{tabular}{cccccc}
\toprule
\begin{tabular}[c]{@{}c@{}}Model/\\ Dataset\end{tabular} &
  \begin{tabular}[c]{@{}c@{}}NLL\\ (Ensemble)\end{tabular} &
  \begin{tabular}[c]{@{}c@{}}NLL\\ (GPED)\end{tabular} &
  \begin{tabular}[c]{@{}c@{}}NLL\\ (EnD\textsuperscript{2})\end{tabular} &
  \begin{tabular}[c]{@{}c@{}}MAE \\ Entropy\\ (GPED)\end{tabular} &
  \begin{tabular}[c]{@{}c@{}}MAE \\Entropy\\ (EnD\textsuperscript{2})\end{tabular} \\ \midrule
\begin{tabular}[c]{@{}c@{}}FCNN/\\ MNIST\end{tabular}  & 0.362 & 0.409 & 0.447 & 0.063 & 0.110 \\ \midrule
\begin{tabular}[c]{@{}c@{}}CNN/\\ MNIST\end{tabular}   & 0.269 & 0.447 & 0.460 & 0.121 & 0.183 \\ \midrule
\begin{tabular}[c]{@{}c@{}}CNN/\\ CIFAR10\end{tabular} & 0.799 & 1.015 & 1.056 & 0.181 & 0.494 \\ \bottomrule
\end{tabular}
}
\end{table} 

\begin{table}[htbp]
\centering
\caption{In-distribution Test set metrics comparison using $U_o$ and student model obtained using the largest width multiplier.}
\label{tab:test-metrics-comparison-uo-large-model}
\resizebox{\linewidth}{!}{
\begin{tabular}{cccccc}
\toprule
\begin{tabular}[c]{@{}c@{}}Model/\\ Dataset\end{tabular} &
  \begin{tabular}[c]{@{}c@{}}NLL\\ (Ensemble)\end{tabular} &
  \begin{tabular}[c]{@{}c@{}}NLL\\ (GPED)\end{tabular} &
  \begin{tabular}[c]{@{}c@{}}NLL\\ (EnD\textsuperscript{2})\end{tabular} &
  \begin{tabular}[c]{@{}c@{}}MAE \\ Entropy\\ (GPED)\end{tabular} &
  \begin{tabular}[c]{@{}c@{}}MAE \\Entropy\\ (EnD\textsuperscript{2})\end{tabular} \\ \midrule
\begin{tabular}[c]{@{}c@{}}FCNN/\\ MNIST\end{tabular}  & 0.362 & 0.401 & 0.408 & 0.069 & 0.099 \\ \midrule
\begin{tabular}[c]{@{}c@{}}CNN/\\ MNIST\end{tabular}   & 0.269 & 0.305 & 0.314 & 0.086 & 0.103 \\ \midrule
\begin{tabular}[c]{@{}c@{}}CNN/\\ CIFAR10\end{tabular} & 0.799 & 0.881 & 0.885 & 0.146 & 0.338 \\ \bottomrule
\end{tabular}
}
\end{table} 

% \columnbreak

\begin{table}[htbp]
\centering
\caption{AUROC for OOD Detection using $U_s$ and student model of the same architecture as the teacher.}
\label{tab:ood-auroc-us-small-model}
\resizebox{\linewidth}{!}{
\begin{tabular}{ccccc}
\toprule
\begin{tabular}[c]{@{}c@{}}Model \& Train Data/\\ OOD Data\end{tabular} & Uncertainty & Ensemble & \begin{tabular}[c]{@{}c@{}}GPED\\ (ours)\end{tabular} & EnD\textsuperscript{2} \\ \midrule
\multirow{2}{*}{\begin{tabular}[c]{@{}c@{}}FCNN-MNIST/\\ KMNIST\end{tabular}} & Total     & 0.929 & 0.875 & 0.761 \\ \cline{2-5} 
                                                                              & Knowledge & 0.948 & 0.926 & 0.861 \\ \midrule
\multirow{2}{*}{\begin{tabular}[c]{@{}c@{}}FCNN-MNIST/\\ notMNIST\end{tabular}} & Total     & 0.944 & 0.736 & 0.659 \\ \cline{2-5} 
                                                                              & Knowledge & 0.990 & 0.803 & 0.720 \\ \midrule

\multirow{2}{*}{\begin{tabular}[c]{@{}c@{}}CNN-MNIST/\\ KMNIST\end{tabular}} & Total     & 0.894 & 0.829 & 0.887 \\ \cline{2-5} 
                                                                              & Knowledge & 0.956 & 0.887 & 0.933 \\ \midrule

\multirow{2}{*}{\begin{tabular}[c]{@{}c@{}}CNN-MNIST/\\ notMNIST\end{tabular}} & Total     & 0.888 & 0.841 & 0.828 \\ \cline{2-5} 
                                                                              & Knowledge & 0.946 & 0.902 & 0.889 \\ \midrule

\multirow{2}{*}{\begin{tabular}[c]{@{}c@{}}CNN-CIFAR10/\\ TIM\end{tabular}}      & Total     & 0.729 & 0.737 & 0.726 \\ \cline{2-5} 
                                                                              & Knowledge & 0.796 & 0.773 & 0.773 \\ \midrule
                                                                              
\multirow{2}{*}{\begin{tabular}[c]{@{}c@{}}CNN-CIFAR10/\\ LSUN\end{tabular}}      & Total     & 0.790 & 0.760 & 0.733 \\ \cline{2-5} 
                                                                              & Knowledge & 0.752 & 0.718 & 0.653 \\ \bottomrule
\end{tabular}
}
\end{table}

\begin{table}[htbp]
\centering
\caption{AUROC for OOD Detection using $U_o$ and student model of the same architecture as the teacher.}
\label{tab:ood-auroc-uo-small-model}
\resizebox{\linewidth}{!}{
\begin{tabular}{ccccc}
\toprule
\begin{tabular}[c]{@{}c@{}}Model \& Train Data/\\ OOD Data\end{tabular} & Uncertainty & Ensemble & \begin{tabular}[c]{@{}c@{}}GPED\\ (ours)\end{tabular} & EnD\textsuperscript{2} \\ \midrule
\multirow{2}{*}{\begin{tabular}[c]{@{}c@{}}FCNN-MNIST/\\ KMNIST\end{tabular}} & Total     & 0.929 & 0.847 & 0.802 \\ \cline{2-5} 
                                                                              & Knowledge & 0.948 & 0.907 & 0.912 \\ \midrule
\multirow{2}{*}{\begin{tabular}[c]{@{}c@{}}FCNN-MNIST/\\ notMNIST\end{tabular}} & Total     & 0.944 & 0.782 & 0.762 \\ \cline{2-5} 
                                                                              & Knowledge & 0.990 & 0.847 & 0.890 \\ \midrule

\multirow{2}{*}{\begin{tabular}[c]{@{}c@{}}CNN-MNIST/\\ KMNIST\end{tabular}} & Total     & 0.894 & 0.829 & 0.884 \\ \cline{2-5} 
                                                                              & Knowledge & 0.956 & 0.890 & 0.937 \\ \midrule

\multirow{2}{*}{\begin{tabular}[c]{@{}c@{}}CNN-MNIST/\\ notMNIST\end{tabular}} & Total     & 0.888 & 0.848 & 0.859 \\ \cline{2-5} 
                                                                              & Knowledge & 0.946 & 0.907 & 0.918 \\ \midrule

\multirow{2}{*}{\begin{tabular}[c]{@{}c@{}}CNN-CIFAR10/\\ TIM\end{tabular}}      & Total     & 0.729 & 0.747 & 0.703 \\ \cline{2-5} 
                                                                              & Knowledge & 0.796 & 0.763 & 0.747 \\ \midrule
                                                                              
\multirow{2}{*}{\begin{tabular}[c]{@{}c@{}}CNN-CIFAR10/\\ LSUN\end{tabular}}      & Total     & 0.790 & 0.759 & 0.728 \\ \cline{2-5} 
                                                                              & Knowledge & 0.752 & 0.736 & 0.650 \\ \bottomrule
\end{tabular}
}
\end{table}

\begin{table}[htbp]
\centering
\caption{AUROC for OOD Detection using $U_o$ and student model obtained by largest width multiplier.}
\label{tab:ood-auroc-uo-large-model}
\resizebox{\linewidth}{!}{
\begin{tabular}{ccccc}
\toprule
\begin{tabular}[c]{@{}c@{}}Model \& Train Data/\\ OOD Data\end{tabular} & Uncertainty & Ensemble & \begin{tabular}[c]{@{}c@{}}GPED\\ (ours)\end{tabular} & EnD\textsuperscript{2} \\ \midrule
\multirow{2}{*}{\begin{tabular}[c]{@{}c@{}}FCNN-MNIST/\\ KMNIST\end{tabular}} & Total     & 0.929 & 0.895 & 0.833 \\ \cline{2-5} 
                                                                              & Knowledge & 0.948 & 0.944 & 0.938 \\ \midrule
\multirow{2}{*}{\begin{tabular}[c]{@{}c@{}}FCNN-MNIST/\\ notMNIST\end{tabular}} & Total     & 0.944 & 0.751 & 0.714 \\ \cline{2-5} 
                                                                              & Knowledge & 0.990 & 0.826 & 0.835 \\ \midrule

\multirow{2}{*}{\begin{tabular}[c]{@{}c@{}}CNN-MNIST/\\ KMNIST\end{tabular}} & Total     & 0.894 & 0.892 & 0.867 \\ \cline{2-5} 
                                                                              & Knowledge & 0.956 & 0.940 & 0.956 \\ \midrule

\multirow{2}{*}{\begin{tabular}[c]{@{}c@{}}CNN-MNIST/\\ notMNIST\end{tabular}} & Total     & 0.888 & 0.887 & 0.856 \\ \cline{2-5} 
                                                                              & Knowledge & 0.946 & 0.941 & 0.935 \\ \midrule

\multirow{2}{*}{\begin{tabular}[c]{@{}c@{}}CNN-CIFAR10/\\ TIM\end{tabular}}      & Total     & 0.729 & 0.750 & 0.705 \\ \cline{2-5} 
                                                                              & Knowledge & 0.796 & 0.798 & 0.783 \\ \midrule
                                                                              
\multirow{2}{*}{\begin{tabular}[c]{@{}c@{}}CNN-CIFAR10/\\ LSUN\end{tabular}}      & Total     & 0.790 & 0.779 & 0.745 \\ \cline{2-5} 
                                                                              & Knowledge & 0.752 & 0.753 & 0.716 \\ \bottomrule
\end{tabular}
}
\end{table}

% \subsection{Supplemental Results for Experiment 5: Ranking based on Uncertainty}

\begin{table}[htbp]
\centering
\caption{nDCG@20 out of 100 randomly selected test inputs using $U_s$ estimator and student model matching the architecture of teacher. Results reported as mean $\pm$ std. dev. over 500 trials.}
\label{tab:ndcg-comparison-us-small-model}
\resizebox{\linewidth}{!}{
\begin{tabular}{cccc}
\toprule
Data                        & Uncertainty & \begin{tabular}[c]{@{}c@{}}GPED\\ (ours)\end{tabular} & EnD\textsuperscript{2}         \\ \midrule
\multirow{2}{*}{FCNN-MNIST} & Total       & 0.947 $\pm$ 0.022                                         & 0.911 $\pm$ 0.031 \\ \cline{2-4} 
                            & Knowledge   & 0.919 $\pm$ 0.030                                          & 0.895 $\pm$ 0.042 \\ \midrule

\multirow{2}{*}{CNN-MNIST} & Total       & 0.852 $\pm$ 0.023                                          & 0.780 $\pm$ 0.040 \\ \cline{2-4} 
                            & Knowledge   & 0.823 $\pm$ 0.038                                          & 0.768 $\pm$ 0.044 \\ \midrule

\multirow{2}{*}{CIFAR10}   & Total       & 0.895 $\pm$ 0.035                                          & 0.853 $\pm$ 0.041 \\ \cline{2-4} 
                            & Knowledge   & 0.825 $\pm$ 0.045                                         & 0.809 $\pm$ 0.051 \\ \bottomrule
\end{tabular}
}
\end{table}

\begin{table}[t]
\centering
\caption{nDCG@20 out of 100 randomly selected test inputs using $U_0$ estimator and student model matching the architecture of teacher. Results reported as mean $\pm$ std. dev. over 500 trials.}
\label{tab:ndcg-comparison-u0-small-model}
\resizebox{\linewidth}{!}{
\begin{tabular}{cccc}
\toprule
Data                        & Uncertainty & \begin{tabular}[c]{@{}c@{}}GPED\\ (ours)\end{tabular} & EnD\textsuperscript{2}         \\ \midrule
\multirow{2}{*}{FCNN-MNIST} & Total       & 0.942 $\pm$ 0.023                                        & 0.907 $\pm$ 0.034 \\ \cline{2-4} 
                            & Knowledge   & 0.922 $\pm$ 0.024                                          & 0.907$\pm$ 0.037 \\ \midrule

\multirow{2}{*}{CNN-MNIST} & Total       & 0.871 $\pm$ 0.050                                         & 0.818 $\pm$ 0.048 \\ \cline{2-4} 
                            & Knowledge   & 0.815 $\pm$ 0.050                                          & 0.747 $\pm$ 0.066 \\ \midrule

\multirow{2}{*}{CIFAR10}   & Total       & 0.903 $\pm$ 0.029                                          & 0.840 $\pm$ 0.047 \\ \cline{2-4} 
                            & Knowledge   & 0.854 $\pm$ 0.041                                         & 0.785 $\pm$ 0.082 \\ \bottomrule
\end{tabular}
}
\end{table}

\begin{table}[t]
\centering
\caption{nDCG@20 out of 100 randomly selected test inputs using $U_0$ estimator and student model obtained using the largest width multiplier. Results reported as mean $\pm$ std. dev. over 500 trials.}
\label{tab:ndcg-comparison-u0-large-model}
\resizebox{\linewidth}{!}{
\begin{tabular}{cccc}
\toprule
Data                        & Uncertainty & \begin{tabular}[c]{@{}c@{}}GPED\\ (ours)\end{tabular} & EnD\textsuperscript{2}         \\ \midrule
\multirow{2}{*}{FCNN-MNIST} & Total       & 0.962 $\pm$ 0.017                                       & 0.940 $\pm$ 0.024 \\ \cline{2-4} 
                            & Knowledge   & 0.953 $\pm$ 0.023                                          & 0.922 $\pm$ 0.020 \\ \midrule

\multirow{2}{*}{CNN-MNIST} & Total       & 0.951 $\pm$ 0.017                                         & 0.908 $\pm$ 0.034 \\ \cline{2-4} 
                            & Knowledge   & 0.920 $\pm$ 0.028                                          & 0.870 $\pm$ 0.036 \\ \midrule

\multirow{2}{*}{CIFAR10}   & Total       & 0.940 $\pm$ 0.021                                         & 0.901 $\pm$ 0.019 \\ \cline{2-4} 
                            & Knowledge   & 0.883 $\pm$ 0.035                                         & 0.883 $\pm$ 0.036 \\ \bottomrule
\end{tabular}
}
\end{table}

%In-distribution test comparisons begin here.

\end{document}